\documentclass[journal]{IEEEtran}
\ifCLASSINFOpdf
\else
\fi
\usepackage{orcidlink}
\usepackage{cite}
\usepackage{algorithmic}
\usepackage{graphicx}
\usepackage{textcomp}
\usepackage{xcolor}
\usepackage{amsmath,amssymb,amsfonts}

\usepackage{textcomp}
\usepackage{hyperref}
\usepackage{amssymb} 
\usepackage{caption} 
\usepackage{booktabs}
\usepackage{subfigure}
\usepackage{multicol,lipsum}
\usepackage{multirow}
\usepackage{algorithm}
\usepackage{romannum}
\usepackage{balance}
\usepackage[misc]{ifsym}
\usepackage{mathrsfs}
\usepackage{stmaryrd} %
\usepackage{amsthm}
\theoremstyle{definition}
\newtheorem{definition}{Definition}

\begin{document}

\title{AutoPRL: Automated Path Representation Learning via a Learnable View Generator}
\author{Sean~Bin~Yang,~\IEEEmembership{Member,~IEEE,}
	Ying~Sun,
	Yunyao~Cheng,
	Hao~Miao,~\IEEEmembership{Member,~IEEE,}
	Jilin~Hu\textsuperscript{*},~\IEEEmembership{Member,~IEEE,}
	Bin~Yang,~\IEEEmembership{Senior Member,~IEEE,}
	Torben Bach Pedersen,~\IEEEmembership{Senior Member,~IEEE,}
	and~Christian~S.~Jensen,~\IEEEmembership{Fellow,~IEEE}
\thanks{Sean Bin Yang, Yunyao Cheng, Hao Miao, Torben Bach Pedersen and Christian S. Jensen are with Aalborg University, Denmark.}
\thanks{Ying Sun is with the Chongqing University of Posts and Telecommunications, China.}%
\thanks{ Jilin Hu and Bin Yang are with East China Normal University, Shanghai, China.}
\thanks{*Corresponding author: Jilin Hu.}
}

%
%

\maketitle
\begin{abstract}
Self-supervised learning has emerged as a powerful paradigm for Path Representation Learning (PRL) in road networks. PRL aims to learn informative representations of paths traversed by moving objects, such as vehicles, where each path consists of a sequence of connected road segments. These representations are essential for various downstream tasks, including travel-time estimation and path ranking. However, existing methods often suffer from ineffective view generation and inefficient training procedures due to their reliance on heuristic augmentations and computationally intensive contrastive learning.
To address these issues, we propose \emph{AutoPRL}, an effective and efficient self-supervised PRL framework that integrates three key components.
The \textit{Automatic Data Augmentation} (\emph{AutoDA}) module employs variational autoencoders to generate semantically meaningful and diverse feature-level path views by modeling each topology-aware node embedding with a Gaussian latent distribution, thereby mitigating information loss and enhancing representation quality.
Second, the \textit{Cross-View Decoupled Contrastive} (\emph{CVDC}) learning mechanism improves training efficiency by disentangling the optimization of positive and negative pairs across node- and path-level views, enabling more stable contrastive learning. Third, an \textit{Adaptive Co-Training} (\emph{ACT}) module dynamically adjusts uncertainty-based loss weights to balance multiple objectives, ensuring efficient and scalable training. 
Extensive experiments on three real-world datasets demonstrate that \emph{AutoPRL} achieves strong and consistent performance in both effectiveness and efficiency across multiple downstream tasks.
\end{abstract}


\IEEEpeerreviewmaketitle

\section{Introduction}
\IEEEPARstart{W}{ith} the rapid proliferation of GPS-enabled devices and ubiquitous mobile computing services, massive volumes of path data are being continuously collected to capture the mobility patterns of vehicles. Effective analysis of such data is fundamental to a broad range of intelligent transportation applications, including travel time estimation \cite{DBLP:journals/pvldb/ZhouSCJK24,DBLP:conf/kdd/YangHGYJ23,DBLP:conf/ijcai/YangGHT021,DBLP:conf/icde/YangGHYTJ22,DBLP:conf/kdd/YangSHXTLYJ26,DBLP:conf/CIKM/TFM}, trajectory mining \cite{DBLP:journals/pvldb/HanCMG22,DBLP:journals/pvldb/ChangTCJQ24,DBLP:conf/icde/0001ZC0X024,DBLP:conf/icde/JiangPRJLW23,DBLP:conf/icde/JensenYGHT24,DBLP:journals/pvldb/GuoXYYKZJ24,DBLP:journals/tkde/XuBL23,DBLP:conf/ijcai/YangMXHWLYJ26}, and path recommendation \cite{DBLP:journals/tkde/YangGY22,DBLP:conf/kdd/YangHGYJ23,DBLP:conf/ijcai/YangGHT021,DBLP:conf/icde/YangGHYTJ22}.
\textit{Path representation learning} (PRL), an essential and popular path data analytics technique, aims at learning a generic path representation that can support various downstream tasks, rather than being tailored to a specific application \cite{DBLP:conf/ijcai/YangGHT021,DBLP:conf/icde/YangGHYTJ22,DBLP:conf/kdd/YangHGYJ23,Path-LLM,MM-Path,DBLP:conf/icde/CURE}.

\begin{figure}[!t]
	\centering
	\includegraphics[scale=0.74]{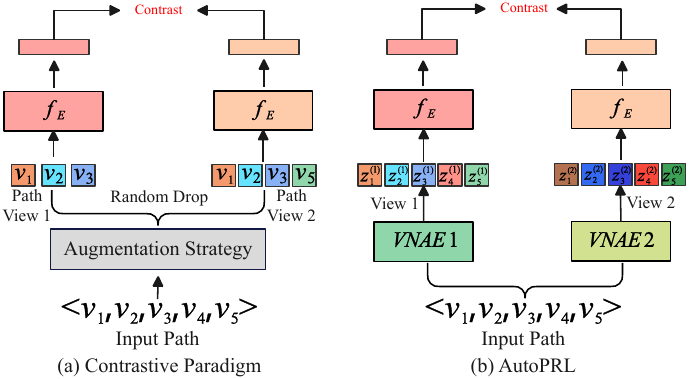}
	\caption{Architectures of the conventional contrastive paradigm and our \emph{AutoPRL} for path representation learning. In (a), heuristic augmentation directly perturbs the discrete path sequence. In (b), \emph{AutoDA} generates feature-level views as continuous embedding sequences from topology-aware node embeddings, rather than discrete node sequences on the road-network graph. Contrast denotes the contrastive loss.}
	\label{fig:LPExample}
	\vspace{-15pt}
\end{figure}
Due to its significance, substantial research efforts have been devoted to developing effective and efficient methods for path representation learning. Among them, contrastive learning-based approaches have garnered considerable attention due to their ability to enhance representation generalization by pulling positive samples closer while pushing negative samples apart in the embedding space \cite{DBLP:journals/vldb/LiuHFLCX23,DBLP:conf/ijcai/YangGHT021,DBLP:conf/icde/YangGHYTJ22,DBLP:conf/www/MaTCZXZCZG24,DBLP:conf/kdd/YangHGYJ23,DBLP:conf/nips/ZhouH00CO24, DBLP:conf/cvpr/HeCXLDG22}. 

More specifically, as illustrated in Figure \ref{fig:LPExample} (a), most existing PRL methods adopt a contrastive learning paradigm comprising three stages:
(1) Given an input path (e.g., $\langle v_1, v_2, v_3, v_4, v_5 \rangle$), augmented views are generated using heuristic, prior-based strategies such as random node drop, resulting in path view 1 ($\langle v_1, v_2, v_3 \rangle$) and path view 2 ($\langle v_1, v_2, v_3, v_5 \rangle$);
(2) These path views are encoded via a shared path encoder to obtain latent representations;
(3) A contrastive loss is applied to align representations of positive pairs while pushing apart negatives. Despite their simplicity, such pre-defined augmentations are inherently limited: they often lead to significant semantic information loss and lack adaptability to the structural characteristics of individual paths. As a result, the learned representations are not only suboptimal but also suffer from poor generalization across tasks and domains. 

Different from existing PRL methods that mainly rely on pre-defined positive samples or manually designed structure-level perturbations, \emph{AutoPRL} adapts learnable view generation to the PRL setting and generates probabilistic feature-level path views. It automatically generates semantically consistent feature-level path views in the latent space based on topology-aware node embeddings derived from the road network. The generated views are continuous embedding sequences rather than discrete topology-level perturbations. In addition, \emph{AutoPRL} adopts a cross-view decoupled contrastive learning mechanism to improve training efficiency and stability, and an adaptive co-training strategy to dynamically balance multiple learning objectives.

These limitations motivate the design of \emph{AutoPRL}, an effective and efficient framework for learning generalizable path representations through an automatic and learnable view generator. \emph{AutoPRL} addresses three key challenges:

\textit{Challenge I: How to effectively construct path views with minimal information loss?} To derive effective and generalizable path representations, existing methods often rely on heuristic, prior-driven view generation strategies. However, such approaches—e.g., random node drop (as illustrated in Figure~\ref{fig:LPExample}(a), where the generated path view retains only a partial segment of the original path)—can result in significant information loss, thereby limiting the representational capacity of the path encoder and leading to suboptimal performance. 

Although recent advances in automatic and learnable view generation have demonstrated effectiveness in graph and vision domains, these techniques are not directly applicable to path representation learning due to the unique spatial dependencies inherent in paths. To this end, as shown in Figure \ref{fig:LPExample}(b), \emph{AutoPRL} generates semantically consistent path views with minimal information loss by leveraging two independent variational autoencoders that explicitly account for spatial dependencies. 
Specifically, it models each topology-aware node embedding with a Gaussian latent distribution and performs sampling from these distributions to generate diverse yet semantically consistent feature-level views for each input path. The generated views are continuous embedding sequences in the latent space rather than discrete paths on the road-network graph. This probabilistic design introduces adaptive feature-level variations without directly modifying the underlying road-network topology. This process constitutes an automatic data augmentation module, referred to as \emph{AutoDA}, which enables \emph{AutoPRL} to learn more effective and generalizable path representations for various downstream tasks.

\textit{Challenge II: How to efficiently conduct contrastive learning for PRL?} Prevailing PRL frameworks commonly adopt \emph{SimCLR}-style contrastive objectives \cite{DBLP:conf/icde/JiangPRJLW23, DBLP:conf/ijcai/YangGHT021}, which rely on a large number of negative samples to achieve discriminative representations. Although effective, these methods incur significant computational and memory overhead due to their dependence on large batch sizes, thereby posing challenges for scalability and efficiency. To facilitate more efficient contrastive training, we adapt the decoupled contrastive learning principle to PRL and develop a \emph{CVDC} learning mechanism that enhances training efficiency by disentangling the optimization of positive and negative sample pairs. By eliminating the competition between positives and negatives within a shared contrastive space, \emph{CVDC} enables more stable and computationally efficient training of the path encoder.

\textit{Challenge III: How to efficiently optimize multi-objective learning in a dynamic setting?} Unlike existing fixed-weight multi-objective optimization strategies, where the contribution of each objective remains static throughout training, path representation learning often involves objectives whose gradients vary dynamically during training. This dynamic behavior necessitates an adaptive optimization mechanism that can effectively balance competing objectives in real time to ensure stable and efficient encoder learning.
To mitigate this, \emph{AutoPRL} incorporates an adaptive loss co-training module that learns uncertainty-based weights for each objective. This dynamic weighting strategy facilitates efficient multi-objective optimization and enhances both the stability and efficiency of path encoder training.

In summary, the \emph{AutoPRL} framework consists of three components: \emph{AutoDA}, \emph{CVDC}, and \emph{ACT} modules. 

First, we introduce \emph{AutoDA}, an automatic and learnable path view generator that adaptively generates semantically meaningful views conditioned on the input path through Gaussian latent modeling, thereby mitigating information loss associated with heuristic augmentations. Second, we propose a \emph{CVDC} mechanism that separately optimizes cross-node and cross-path objectives. This decoupling strategy allows the encoder to simultaneously capture local structural dependencies among nodes and global semantic patterns across entire paths. By decoupling the competition between positive and negative samples in the shared contrastive space, \emph{CVDC} facilitates more stable optimization and enhances computational efficiency in path encoder training. Finally, the \emph{ACT} module dynamically balances multiple learning objectives via uncertainty-based weighting, enhancing both training efficiency and model generalization.

The main contributions of this paper are summarized as follows:
\begin{itemize}
	\item We propose \emph{AutoPRL}, an automated path representation learning framework that adapts learnable view generation to PRL. Compared with previous PRL methods that mainly rely on pre-defined positive samples or heuristic augmentations, \emph{AutoPRL} generates semantically consistent feature-level path views through Gaussian latent modeling, reducing information loss and improving generalization.
	\item We adapt decoupled contrastive learning to PRL and formulate a cross-view contrastive learning component with cross-path and cross-node objectives. This component captures both global path-level semantics and local node-level consistency while improving training efficiency.%
	\item We incorporate an adaptive co-training strategy to jointly train the view generator and path encoder in an end-to-end manner. This strategy introduces uncertainty-based adaptive weighting into PRL to dynamically balance multiple loss components, reducing the need for manual weight tuning and improving training stability.
	\item We conduct extensive experiments on three real-world datasets across multiple downstream tasks to assess the effectiveness and generality of \emph{AutoPRL}.
\end{itemize}

\section{Related Work}
\label{sec:RW}
\subsection{Trajectory Representation Learning}
With the remarkable success of self-supervised learning, the pursuit of generic trajectory representations has received increasing attention, as such representations can support a wide range of downstream applications in urban computing.
A variety of methods have been proposed to learn vector representations of trajectories. Existing trajectory representation learning (TRL) approaches can be broadly categorized into spatial trajectory–based (i.e., path-based) methods and spatio-temporal trajectory–based methods.

\noindent
\textbf{Spatial Trajectory-based Methods.} Spatial Trajectory Representation Learning (STRL), also known as path representation learning (PRL), focuses solely on spatial information to derive a generic path representation. In this setting, existing works are primarily divided into two classes, recurrent neural network (RNN) based methods and Transformer based methods. \emph{PIM} \cite{DBLP:conf/ijcai/YangGHT021} proposes a first RNN-based mutual information maximization framework to learn generic path representations through curriculum negative sampling. 
To improve upon this, \emph{TPR} \cite{DBLP:conf/icde/YangGHYTJ22} treats temporal information as weak labels to construct diverse contrastive samples and proposes a weakly supervised contrastive learning method with a curriculum learning strategy based on LSTMs. However, RNN-based methods struggle to capture long-range path dependencies and generally achieve significantly lower accuracy than Transformer-based methods.
Building on this, \emph{LightPath} \cite{DBLP:conf/kdd/YangHGYJ23} proposes a scalable and lightweight path representation learning framework that employs varying mask ratios for path augmentation and adopts relational reasoning within a Transformer architecture to learn generic path representations.
More recently, \emph{MM-Path} \cite{MM-Path} and \emph{Path-LLM} \cite{Path-LLM} extend PRL into the multi-modality domain by incorporating image and textual modalities, respectively. However, these PRL methods still rely on pre-defined augmentation strategies, which limits their generalization ability.

\noindent
\textbf{Trajectory-based Methods.} In contrast, trajectory representation learning (TRL) aims to learn generic representations by jointly modeling spatial and temporal information, typically through Transformer architectures. \emph{Toast} \cite{DBLP:conf/cikm/ChenLCBLLCE21} introduces a traffic context–aware skip-gram module and a trajectory-enhanced Transformer module to improve trajectory representation learning. \emph{START} \cite{DBLP:conf/icde/JiangPRJLW23} incorporates temporal regularities and travel semantics via a random masking strategy and further enhances learning with data augmentation techniques such as trajectory trimming, temporal shifting, and contrastive learning. \emph{TrajRL} \cite{DBLP:conf/icde/XiaZCCYD25} proposes a self-supervised trajectory representation learning model that captures multi-scale spatio-temporal correlations from both road segments and segment sequences. It further incorporates multi-faceted temporal information, including time intervals, to model context-aware temporal patterns within trajectories. \emph{RED} \cite{DBLP:journals/pvldb/ZhouSCJK24} leverages comprehensive trajectory information and adopts a road-aware masking strategy to learn more effective trajectory representations. We adopt \emph{LightPath,} \emph{Toast}, \emph{START}, \emph{TrajRL} and \emph{RED} as baselines since they are Transformer-based methods, and additionally include \emph{PIM} as a baseline because it represents a classical path representation learning approach.

\subsection{Contrastive Learning}

In recent years, contrastive learning has made significant strides in unsupervised and self-supervised learning. Its core objective is to maximize similarity between positive pairs while minimizing similarity between negative pairs, using augmented views of the same data. Several frameworks have been proposed, including \emph{SimCLR} \cite{DBLP:conf/cvpr/ChenH21}, \emph{BYOL} \cite{DBLP:conf/nips/GrillSATRBDPGAP20}, and \emph{MoCo} \cite{DBLP:conf/cvpr/He0WXG20}.
However, a key challenge remains: model collapse, where embeddings degenerate into constant values, undermining feature learning. To address this, two main strategies have emerged. The first relies on negative samples, which requires large batch sizes and extensive GPU memory (e.g., \emph{PIM} \cite{DBLP:conf/ijcai/YangGHT021}). The second uses momentum encoders with stop-gradient and sliding average mechanisms to preserve representational diversity. Studies suggest that stop-gradient, rather than the sliding average, plays a central role in preventing collapse.
\emph{SimSiam} \cite{DBLP:conf/cvpr/ChenH21}, and its variant \emph{LightPath} \cite{DBLP:conf/kdd/YangHGYJ23}, leverage stop-gradient with projection layers, achieving strong performance without large batch sizes or extensive negative sampling. This makes \emph{SimSiam}-style approaches efficient for downstream tasks. Nonetheless, contrastive learning methods still face challenges in scalability and resistance to collapse, warranting further research into more efficient and effective solutions.

\section{Preliminaries}
\label{sec:preli}
We proceed to present the necessary preliminaries and then define the problem addressed. 
\subsection{Basic Definition}

\begin{definition}[\textbf{Road Network}] A road network is represented as a graph $\mathcal{G} = (\mathcal{V}, \mathcal{E})$, where $\mathcal{V}$ is a set of nodes $v_{i}$, each representing a road intersection, and $\mathcal{E} \subseteq \mathcal{V} \times \mathcal{V}$ is a set of edges $e_{i}$, with each edge $e_i=(v_i,v_j)$ denoting a directed road segment from node $v_i$ to $v_j$.
\end{definition}

\begin{definition}[\textbf{Path}] A path $p=\langle v_1,v_2,\cdots,v_n\rangle$ is defined as a sequence of connected nodes, where each $v_i \in \mathcal{V}$ represents a node in the path. 
Specifically, for each adjacent pair of edges, $e_i=(v_i,v_j)$ and $e_{i+1}=(v_j,v_k)$ share a common node $v_j$, where $i,j,k \leq n$.
\end{definition}

\begin{definition}[\textbf{Node Embedding Vector}] In a road network graph $\mathcal{G}$, the embedding of a node is denoted by an embedding vector $h_n \in \mathbb{R}^{d_n}$, where $d_n$ represents the dimensionality of the node embedding. This vector is typically obtained using graph representation learning models such as \emph{Node2vec} \cite{DBLP:conf/kdd/GroverL16}.
\end{definition}

\begin{definition}[\textbf{Path View}]
	Given a path $p_i=\langle v_1,v_2,\cdots,v_n\rangle$, a \emph{path view} refers to an augmented representation of $p_i$ used as a positive sample in contrastive learning. Depending on the augmentation mechanism, a path view can be either a discrete node sequence generated by heuristic structure-level augmentation (e.g., random node dropping) or a continuous node-embedding sequence generated by feature-level augmentation. In this work, \emph{AutoDA} generates the latter type of path view in the latent embedding space.
\end{definition}
\vspace{-10pt}
\begin{figure}[htp]
	\centering
	\includegraphics[scale=0.5]{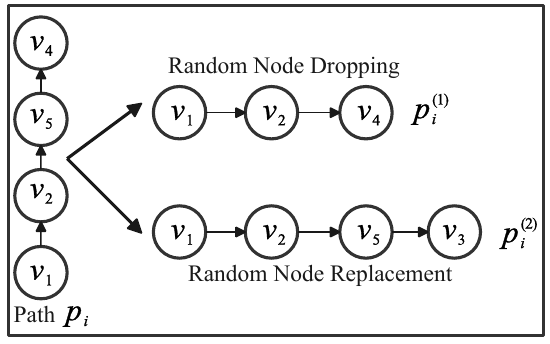}
\caption{An example of heuristic path view generation.}
	\label{fig:dg}
	\vspace{-10pt}
\end{figure}

\subsection{Problem Definition}
Given a set of paths $\mathcal{P}=\bigl\{p_1,p_2,\cdots,p_{|\mathcal{P}|}\bigr\}$ and a road network graph $\mathcal{G}$, \emph{AutoPRL} aims to learn a generic embedding vector, which can be defined as $\mathit{PE}_{\psi}(p_i) \rightarrow \mathbb{R}^{d_p}$ for each path $p_i \in \mathcal{P}$, where $d_p$ is the dimensionality of the path embedding vector.

\subsection{Module Background}

\noindent
\textbf{Transformer Module.}
Given an input path token sequence represented by matrix $Z \in \mathbb{R}^{n \times d_{n}}$, where $n$ is the length of the sequence, and $d_{n}$ is the dimensionality of each embedding vector, the Transformer module starts by adding a position embedding to $\mathit{Z}$ to preserve the order of tokens. It then transforms $\mathit{Z}$ to three distinct matrices: the query matrix $\mathit{Q}=\mathit{ZW}^{\mathit{Q}}\in \mathbb{R}^{n \times d_{n}}$, the key matrix $\mathit{K}=\mathit{ZW}^{\mathit{K}} \in \mathbb{R}^{n \times d_{n}}$, and the value matrix $\mathit{V}=\mathit{ZW}^{\mathit{V}} \in \mathbb{R}^{n \times d_{n}}$, where $\mathit{W}^{\mathit{Q}} $,  $\mathit{W}^{\mathit{K}} $, and $\mathit{W}^{\mathit{V}}$ are learnable weight matrices in $\mathbb{R}^{d_n \times d_{n}}$. The self-attention mechanism computes the attention correlation matrix as
\vspace{-5pt}
\begin{equation}
	\label{eq:trans}
	\operatorname{Attention}(\mathit{Q, K, V})=\operatorname{softmax}\left(\frac{\mathit{Q K}^{\top}}{\sqrt{d_n}}\right) V \in \mathbb{R}^{n \times d_n}
\end{equation}

\noindent
where each element $\alpha_{j,k}$ in the softmax matrix reflects the attention score between $j$-th and $k$-th tokens in the matrix $\mathit{Z}$. 

\subsection{Solution Overview}
\begin{figure}[htp]
	\centering
	\includegraphics[scale=0.68]{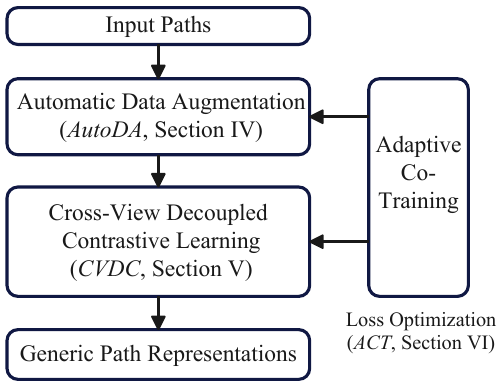}
	\caption{Solution Overview.} 
	\label{fig:so}
	\vspace{-10pt}
\end{figure}

Figure \ref{fig:so} shows an overview of the proposed \emph{AutoPRL}, which consists of three modules: (1) \emph{AutoDA}, (2) \emph{CVDC} Learning, and (3) \emph{ACT}. The details of these modules are provided in Sections \ref{sec:AutoDA}, \ref{sec:CVDC}, and \ref{sec:ACT}.

\section{AutoDA}
\label{sec:AutoDA}
\begin{figure*}[t]
	\centering
	\includegraphics[scale=0.52]{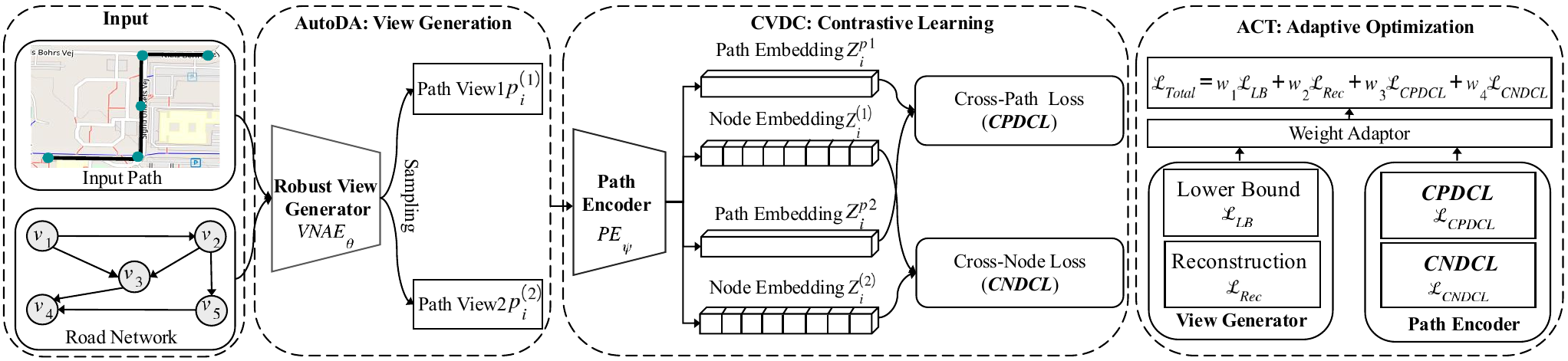}
	\caption{Overall architecture of \emph{AutoPRL}. The framework consists of three modules: \emph{AutoDA} for automatic path view generation, \emph{CVDC} for contrastive representation learning with cross-path and cross-node objectives, and \emph{ACT} for adaptive multi-objective optimization.}

	\label{fig:overview}
	\vspace{-10pt}
\end{figure*}

Figure~\ref{fig:overview} provides an overview of the \emph{AutoPRL} framework, which consists of three modules: \emph{AutoDA} for view generation, \emph{CVDC} for contrastive representation learning, and \emph{ACT} for adaptive optimization.
	
We first detail the \emph{AutoDA} module. Given an input path and the corresponding road-network context, \emph{AutoDA} first maps the path into topology-aware spatial embeddings. It then employs two Gaussian-distribution-based VNAEs to model sequential correlations among nodes and generate two semantically consistent feature-level path views. These generated views are subsequently used by the \emph{CVDC} module for contrastive learning. The intuition is that path augmentation should be dynamic and data-driven, rather than relying on static heuristic transformations.

\subsection{Spatial Embedding} 
\begin{figure}[!t]
	\centering
	\includegraphics[scale=0.42]{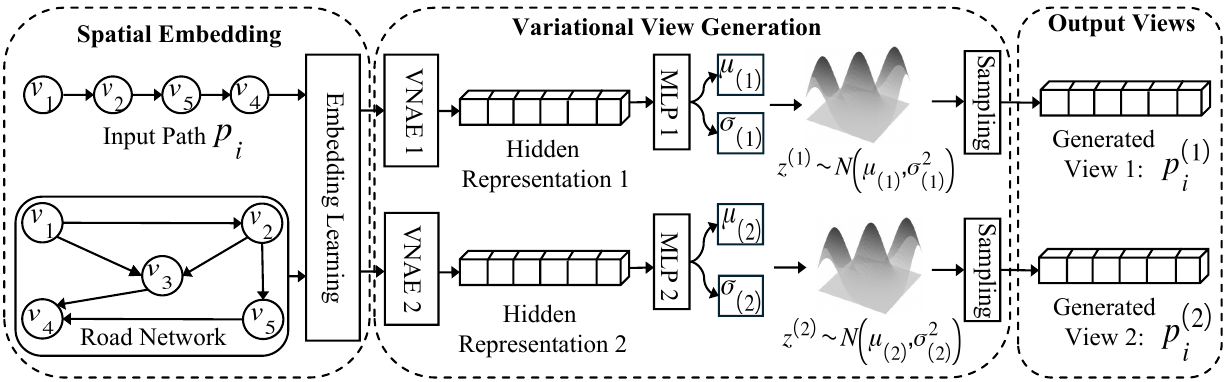}
\caption{Automatic path view generation in \emph{AutoDA}. The input path is first mapped into topology-aware spatial embeddings. Two VNAEs then encode the path into hidden representations, from which MLPs parameterize Gaussian latent distributions. By sampling from these distributions, \emph{AutoDA} generates two semantically consistent feature-level views, $p_i^{(1)}$ and $p_i^{(2)}$.}
	\label{fig:autoda}
\end{figure}

As shown in Figure \ref{fig:autoda}, topology-constrained spatial embedding is fundamental to constructing effective initial path representation vectors. By incorporating the structural properties of the road network, this approach ensures that each node embedding is informed by its local topology, promoting semantic consistency among connected nodes and encouraging similar embeddings for spatially and functionally related segments. 

Motivated by recent advances in graph embedding learning \cite{DBLP:conf/kdd/GroverL16,DBLP:journals/pvldb/YeLC24}, we aim to learn high-quality node embedding vectors that effectively capture the structural properties of road networks. Given a road network represented as a graph $\mathcal{G} = (\mathcal{V}, \mathcal{E})$, where $\mathcal{V}$ denotes intersections and $\mathcal{E}$ denotes road segments, we adopt the \emph{Node2vec} algorithm \cite{DBLP:conf/kdd/GroverL16} to encode topological relationships among nodes. \emph{Node2vec} simulates biased random walks to explore local and global neighborhood structures, and learns embedding vectors by preserving node co-occurrence patterns in a low-dimensional latent space, which can be defined as:
\begin{equation}
	\vec{\mathit{Z}}_{\mathcal{G}}=\operatorname{Node2vec}(\mathcal{G}(\mathcal{V},\mathcal{E}))
\end{equation}
\noindent
where $\vec{\mathit{Z}}_{\mathcal{G}}=\{\vec{z}_v \mid v\in\mathcal{V}\}$ denotes the topology-aware node embedding matrix learned from the road network, and $\vec{z}_v\in\mathbb{R}^{d_{\mathcal{G}}}$ is the embedding of node $v$.
Given a path $p=\langle v_1,v_2,v_5,v_4\rangle \in \mathcal{P}$, the corresponding initial path node sequence embedding is defined as $\vec{\mathit{Z}}_{p} = \{\vec{z}_1, \vec{z}_2, \vec{z}_5, \vec{z}_4\}$, where each $\vec{z}_i $ corresponds to the embedding of node $v_i$ and $\vec{\mathit{Z}}_\mathit{p}$ denotes the set of path sequence embeddings.

\subsection{Automatic Data Augmentation}
To facilitate efficient and automatic path representation learning, we introduce an automatic and data-driven framework for generating contrastive views. Unlike existing methods that rely on manually crafted rules or domain-specific heuristics, our approach learns feature-level augmentation patterns directly from topology-aware node embeddings.

At the core of our method is the \emph{AutoDA} module, which employs a variational node autoencoder as a generative model for producing diverse and semantically meaningful feature-level path views based on learned Gaussian distributions. These views retain information encoded in the input embeddings while providing sufficient variability for contrastive learning.

\paragraph{Variational Node Encoder.}
Inspired by \emph{AutoST} \cite{DBLP:conf/www/ZhangHXWLY23}, the \emph{AutoDA} module adopts two distinct Variational Node AutoEncoders (VNAEs), i.e., variational LSTMs \cite{DBLP:journals/csur/LiangPLYR24}, to generate two augmented path views. The two VNAEs have the same architecture but use non-shared parameters, enabling them to capture different latent variations of the same input path.

Given the initial path embedding sequence $\vec{\mathit{Z}}_{p_i}$, we feed it into two VNAE encoders to obtain two latent hidden representations:

\begin{equation}
	\begin{aligned}
		\vec{\mathit{H}}^{(1)} &= \mathit{VNAE}_1(\vec{\mathit{Z}}_{p_i},\theta_{1}), \\
		\vec{\mathit{H}}^{(2)} &= \mathit{VNAE}_2(\vec{\mathit{Z}}_{p_i},\theta_{2}),
	\end{aligned}
\end{equation}
\noindent
where $\theta_1$ and $\theta_2$ denote the learnable parameters of the two VNAE encoders, and $\vec{\mathit{H}}^{(1)}$ and $\vec{\mathit{H}}^{(2)}$ denote the corresponding hidden node representations.

To model the latent node distribution of each view, we use an MLP to parameterize a Gaussian distribution from the hidden representation $\vec{\mathit{H}}^{(k)}$, where $k\in\{1,2\}$. Specifically, the MLP predicts the mean vector $\mu_{(k)}$ and standard-deviation vector $\sigma_{(k)}$, yielding the node-level approximate posterior
$q_{\phi}(\mathit{Z}^{(k)}\mid\mathit{H}^{(k)})
=\mathcal{N}(\mu_{(k)},\operatorname{diag}(\sigma_{(k)}^{2}))$.
We then sample node-level latent embeddings from this distribution to generate diverse yet semantically consistent augmented path views.

\paragraph{Rationale of VAE-based View Generation.}
The VNAE module is not used merely as an encoding architecture, but as a probabilistic view generator for path augmentation. Compared with deterministic encoders, the VNAE learns a continuous latent distribution and generates augmented path views through latent sampling. This property is particularly suitable for self-supervised PRL, where the diversity and semantic consistency of contrastive views directly affect the quality of learned representations.
We adopt Gaussian latent distributions because they provide a tractable and stable way to model the uncertainty of road-network paths. Paths traversed by moving objects may exhibit stochastic variations caused by route choices, traffic conditions, and local road-network structures, while still preserving their underlying route semantics. The Gaussian latent space enables efficient reparameterized sampling and stable optimization, allowing the model to generate diverse yet semantically consistent path views.

\paragraph{Automatic View Generation} The decoders of the two VNAE modules produce two semantically diverse node-level augmentations, denoted as $p_i^{(1)}$ and $p_i^{(2)}$. 
Specifically, $p_i^{(1)}$ and $p_i^{(2)}$ are continuous node-embedding sequences in the latent space, rather than discrete node sequences mapped back to the original road-network graph. Thus, \emph{AutoDA} performs feature-level augmentation on topology-aware node embeddings instead of explicit structure-level augmentation. Here, ``topology-aware'' refers to the road-network structural information encoded in the input node embeddings and does not imply that \emph{AutoDA} generates topologically valid discrete path perturbations.

In contrast to recent path contrastive learning approaches that rely on heuristic or stochastic augmentation techniques—such as path node masking or dropping \cite{DBLP:conf/kdd/YangHGYJ23,DBLP:conf/icde/JiangPRJLW23}—\emph{AutoDA} introduces a principled, learnable augmentation mechanism by modeling sequential dependencies directly within the data. 
This adaptive and data-driven strategy generates augmented views that preserve the underlying semantic structure while introducing meaningful variation, thereby enhancing spatial path representation. As a result, \emph{AutoDA} offers task-irrelevant, data-driven self-supervised signals that improve the effectiveness, expressiveness, and generalizability of the learned path representations.

\subsection{Objective Optimization for Augmentation} To effectively train the learnable VNAEs, we introduce two objectives that constrain the latent variable $\mathit{Z}$ during the learning process, aiming to align the distribution of reconstructed representations with that of the original input embedding $\mathit{H}$.
We optimize the variational lower bound $\mathscr{L}_{\mathit{LB}}$ using gradient-based learning. Since direct sampling from the approximate posterior introduces stochasticity into the latent variable, we adopt the reparameterization trick to enable gradient propagation through the sampling process. Specifically, for a node-level latent variable with approximate posterior
$q_{\phi}(\mathit{Z}\mid\mathit{H})
=\mathcal{N}(\mu,\operatorname{diag}(\sigma^{2}))$,
we rewrite the sampling operation as
\begin{equation}
	\mathit{Z}=\mu+\sigma\odot\varepsilon,
	\qquad
	\varepsilon\sim\mathcal{N}(0,I),
\end{equation}
where $\mu$ and $\sigma$ denote the mean and standard deviation of the approximate posterior, respectively. This formulation separates the stochastic component $\varepsilon$ from the learnable distribution parameters, thereby enabling end-to-end gradient-based optimization.

We apply a denoising autoencoding criterion to the VNAEs by introducing corrupted input with Gaussian noise,
$\tilde{z}=z+\varepsilon_{\text{noise}}$,
where
$\varepsilon_{\text{noise}}\sim\mathcal{N}(0,\sigma_{\text{noise}}^{2}I)$.

\begin{equation}
	\mathscr{L}_{\mathit{LB}}=\mathscr{L}_\mathit{LB}^{(1)}+\mathscr{L}_\mathit{LB}^{(2)}
\end{equation}
\noindent
In particular:

\begin{equation}
	\begin{aligned}
		\mathscr{L}_{\mathit{LB}}^{(1)}
		=&-D_{\mathit{KL}}\left(
		q_\phi\left(\mathit{Z}^{(1)} \mid \mathit{H}^{(1)}\right)
		\| p_s\left(\mathit{Z}^{(1)}\right)
		\right) \\
		&+\mathbb{E}_{q_\phi\left(\mathit{Z}^{(1)} \mid \mathit{H}^{(1)}\right)}
		\left[
		\log p_\theta\left(\mathit{H}^{(1)} \mid \mathit{Z}^{(1)}\right)
		\right] \\
		\mathscr{L}_{\mathit{LB}}^{(2)}
		=&-D_{\mathit{KL}}\left(
		q_\phi\left(\mathit{Z}^{(2)} \mid \mathit{H}^{(2)}\right)
		\| p_s\left(\mathit{Z}^{(2)}\right)
		\right) \\
		&+\mathbb{E}_{q_\phi\left(\mathit{Z}^{(2)} \mid \mathit{H}^{(2)}\right)}
		\left[
		\log p_\theta\left(\mathit{H}^{(2)} \mid \mathit{Z}^{(2)}\right)
		\right].
	\end{aligned}
	\label{eq:ldp}
\end{equation}
\noindent
where, for $k\in\{1,2\}$, the first term is the KL regularization term that aligns the approximate posterior
$q_{\phi}(\mathit{Z}^{(k)}\mid\mathit{H}^{(k)})$
with the progress-based prior
$p_s(\mathit{Z}^{(k)})$,
and the second term is the reconstruction likelihood that encourages the latent representation to preserve the semantic and structural information encoded in the input embeddings $\mathit{H}^{(k)}$.

We then define the reconstruction objective as $\mathscr{L}_{\mathit{Rec}}(\mathscr{L}_{\mathit{Rec}}^{(1)}, \mathscr{L}_{\mathit{Rec}}^{(2)})$, which evaluates how well the latent variable $\mathit{Z}_{(\cdot)}$ preserves the semantic and topology-aware information encoded in the input node embeddings.

\begin{equation}
	\mathscr{L}_{\mathit{Rec}} = \mathscr{L}_{\mathit{Rec}}^{(1)} +\mathscr{L}_{\mathit{Rec}}^{(2)}
\end{equation}

\noindent
where
\begin{equation}
	\begin{aligned}
		\mathscr{L}_{\mathit{Rec}}^{(1)} &=\frac{1}{N}\sum_{i=1}^{N}\sum_{j=1}^{M}(\mathit{H}_{ij}^{(1)}-\mathit{H}_{ij\prime}^{{(1)}})^{2} \\
		\mathscr{L}_{\mathit{Rec}}^{(2)} &=\frac{1}{N}\sum_{i=1}^{N}\sum_{j=1}^{M}(\mathit{H}_{ij}^{(2)}-\mathit{H}_{ij\prime}^{{(2)}})^{2}
	\end{aligned}
\end{equation}

\noindent
where $\mathit{H}_{ij}^{(k)}$ and $\mathit{H}_{ij\prime}^{(k)}$ represent the original and reconstructed node embeddings of the $j$-th node within path $i$, where $k \in 	\{1,2\}$.

\begin{figure}[!t]
	\centering
	\includegraphics[scale=0.51]{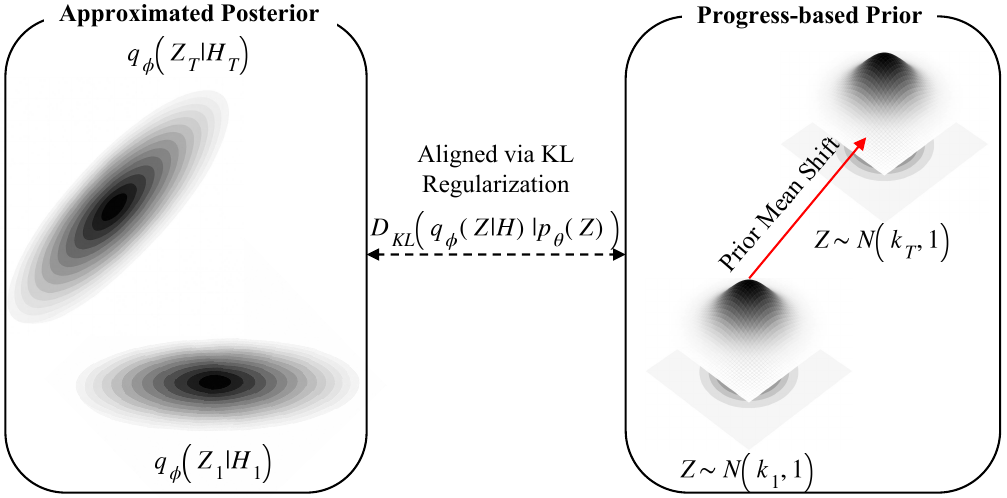}
	\caption{Illustration of the progress-based prior. The approximate posterior is aligned with a dynamic Gaussian prior through KL regularization. During training, the mean of the prior gradually shifts from $\boldsymbol{\mu}_{s_1}$ to $\boldsymbol{\mu}_{s_T}$, which improves the alignment between the prior and the evolving latent space.}
	\vspace{-10pt}
	\label{fig:progre}
\end{figure}

\paragraph{Progress-based Prior.}
Moreover, we introduce a progress-based prior distribution for the latent
variables. Instead of using a fixed standard normal prior
$\mathcal{N}(\mathbf{0},I)$ throughout training, we employ a dynamic
Gaussian prior whose mean evolves with the training progress. Specifically,
at training stage $s$, the prior is defined as
\begin{equation}
	p_s(\mathit{Z})
	=
	\mathcal{N}(\boldsymbol{\mu}_s,I),
\end{equation}
where $\boldsymbol{\mu}_s$ denotes the progress-dependent prior mean and
$I$ is the identity covariance matrix.

The progress-based prior serves as an adaptive regularization strategy for
variational learning. A fixed standard normal prior imposes the same latent
regularization throughout training, which may be restrictive while the
latent representation space is still evolving. In contrast, we gradually
adjust the prior mean from $\boldsymbol{\mu}_{s_1}$ at the initial stage to
$\boldsymbol{\mu}_{s_T}$ at the final stage. This progressive adjustment
allows the prior to better follow the evolving latent space and provides
smoother KL regularization during optimization.

For the learned approximate posterior, we write
\begin{equation}
	q_{\phi}(\mathit{Z}\mid\mathit{H})
	=
	\mathcal{N}
	\left(
	\boldsymbol{\mu}_{Z},
	\boldsymbol{\Sigma}_{Z}
	\right),
	\qquad
	\boldsymbol{\Sigma}_{Z}
	=
	\operatorname{diag}
	\left(
	\boldsymbol{\sigma}_{Z}^{2}
	\right),
\end{equation}
where $\boldsymbol{\mu}_{Z}$ and $\boldsymbol{\sigma}_{Z}$ denote the
mean and standard-deviation vectors of the approximate posterior,
respectively.

The KL regularization term can therefore be written as
\begin{equation}
	\begin{aligned}
		\mathscr{L}_{\mathit{KL}}
		&=
		D_{\mathit{KL}}
		\left(
		q_{\phi}(\mathit{Z}\mid\mathit{H})
		\|
		p_s(\mathit{Z})
		\right) \\
		&=
		D_{\mathit{KL}}
		\left(
		\mathcal{N}
		(\boldsymbol{\mu}_{Z},\boldsymbol{\Sigma}_{Z})
		\|
		\mathcal{N}
		(\boldsymbol{\mu}_{s},I)
		\right) \\
		&=
		\frac{1}{2}
		\left[
		\operatorname{tr}(\boldsymbol{\Sigma}_{Z})
		+
		(\boldsymbol{\mu}_{s}-\boldsymbol{\mu}_{Z})^{T}
		(\boldsymbol{\mu}_{s}-\boldsymbol{\mu}_{Z})
		-
		D
		-
		\log
		\left|
		\boldsymbol{\Sigma}_{Z}
		\right|
		\right],
	\end{aligned}
\end{equation}
where $D$ denotes the latent dimensionality and
$\operatorname{tr}(\cdot)$ denotes the matrix trace.

\section{Cross-View Decoupled Contrastive Learning}
\label{sec:CVDC}
To ensure the learning of generic path representations that generalize across diverse downstream applications, we adopt decoupled contrastive learning to define the overall training objectives. As illustrated in Figure~\ref{fig:overview}, given the two distinct path views generated by the \emph{AutoDA} module, we first describe the path encoder and then present the formulation of the cross-path and cross-node decoupled contrastive losses. The key intuition is that contrastive learning should enable efficient training while maintaining strong performance on downstream tasks.

\subsection{Path Encoder}
To further capture the semantic information of a given path, we employ a Transformer-based path encoder, as defined in Eq.~\ref{eq:trans}. Formally, as shown in Figure \ref{fig:overview}, given two distinct node sequence embeddings for a path $p_i$, denoted as $p_i^{(1)}=\langle \vec{z}^{(1)}_1,\vec{z}^{(1)}_2,\cdots, \vec{z}^{(1)}_{|p_i|} \rangle$  and $p_i^{(2)}=\langle \vec{z}^{(2)}_1,\vec{z}^{(2)}_2,\cdots, \vec{z}^{(2)}_{|p_i|} \rangle$, generated from the learnable view generator, we first enrich both sequences with positional encodings to preserve the order of nodes.
These position-enhanced sequences are then fed into the Transformer encoder to produce high-level node-level embeddings, denoted as $\vec{p}_i^{(1)}=\text{Transformer}(p_i^{(1)})$ and $\vec{p}_i^{(2)}=\text{Transformer}(p_i^{(2)})$ . To obtain the final path-level embeddings, we apply an aggregation function over the transformed edge embeddings in each view, resulting in two path embeddings, $\mathit{Z}_i^{p1}=\vec{\mathit{H}}_{\vec{p}_i}^{(1)} \in \mathbb{R}^{d_p}$ and $\mathit{Z}_i^{p2}=\vec{\mathit{H}}_{\vec{p}_i}^{(2)} \in \mathbb{R}^{d_p}$.

\subsection{Cross-Path Decoupled Contrastive Learning}
\label{section:global}
Given a batch of $N$ samples (e.g., paths), $\bigl\{p_1,p_2,\cdots,p_N\bigr\}$, 
let $\vec{\mathit{H}}_{\vec{p}_i}^{(1)}$  and $\vec{\mathit{H}}_{\vec{p}_i}^{(2)}$ denote two embeddings of the generated view $p_i^{(1)}$ and $p_i^{(2)}$  of the $i$-th path $p_i$. We define the complete set of generated views in the batch as $B=\{p_i^{(k)}| k \in \{1,2\}, i \in [1, N ]\}$, resulting in a total of $2N$ views. Following the \emph{SimCLR} framework \cite{DBLP:conf/cvpr/ChenH21}, contrastive learning is formulated as a classification task for each generated view $p_i^{(k)}$, where the objective is to correctly identify its corresponding positive pair $p_i^{(u)}$ (with $u \neq k$) among all other views in $B$. All remaining $2N - 2$ views are treated as negative samples. 
For each view $p_i^{(k)}$, we use a cross-entropy objective to compute the contrastive loss $\mathscr{L}_i^{(k)}$. The objective pulls the positive pair, i.e., two views generated from the same path, closer in the embedding space, while pushing negative pairs, i.e., views generated from different paths, apart. The overall \emph{SimCLR} loss across the batch is $\mathscr{L}_{\mathit{SimCLR}}=\sum_{k \in \{1,2\}, i \in [1,N]}\mathscr{L}_i^{(k)}$.

\vspace{-10pt}
\begin{align}
	\small
	\mathscr{L}_i^{(k)} 
	&= -\left(\left\langle\vec{H}_{\vec{p}_i}^{(1)}, \vec{H}_{\vec{p}_i}^{(2)}\right\rangle / t \right) 
	+ \log \Bigg(
	\exp \left(\left\langle \vec{\mathit{H}}_{\vec{p}_i}^{(1)}, \vec{\mathit{H}}_{\vec{p}_i}^{(2)} \right\rangle / t \right) \notag \\
	&\quad + \sum_{\substack{j=1 \\ j \neq i}}^{N} \sum_{l=1}^{2}
	\exp \left( \left\langle \vec{\mathit{H}}_{\vec{p}_i}^{(k)}, \vec{\mathit{H}}_{\vec{p}_j}^{(l)} \right\rangle / t \right) \Bigg) \tag{9}
	\label{eq:simclr}
\end{align}
\noindent
where $\exp(\cdot)$ denotes the exponential operation, $\langle\cdot,\cdot\rangle$ denotes the inner product, and $t$ is the temperature parameter.

However, the self-supervised contrastive learning objective in Eq.~\ref{eq:simclr} relies on a large batch size to provide a sufficiently diverse set of negative samples. When the batch size is small, the training process suffers from significant negative-positive coupling (NPC) fluctuations, which can degrade representation quality and training stability. Motivated by decoupled contrastive learning techniques \cite{DBLP:conf/eccv/YehHHLCL22}, we argue that effective and efficient path embedding learning in real-world scenarios should support contrastive learning under small-batch settings. This requires a learning strategy that can mitigate NPC effects while preserving discriminative power in low-sample regimes.

In this paper, we adapt the decoupled contrastive learning principle to path-level representation learning and formulate a \textit{Cross-Path Decoupled Contrastive loss}  (\emph{CPDCL})  to better capture global semantic information (i.e., path-level) while addressing the limitations of conventional contrastive learning in small-batch settings. Specifically, \emph{CPDCL} removes the positive pair term---i.e., 
$\exp \left(\left\langle \vec{\mathit{H}}_{\vec{p}_i}^{(1)}, 
\vec{\mathit{H}}_{\vec{p}_i}^{(2)} \right\rangle / t \right)$---from the denominator of the standard \emph{SimCLR} loss (Eq.~\ref{eq:simclr}), thereby decoupling the gradient flow between positive and negative pairs. As illustrated in Figure~\ref{fig:overview}, for each path $p_i$ in the batch, the overall \emph{CPDCL} objective is $\mathscr{L}_{\mathit{CPDCL}}= \sum_{k \in\{1,2\}, i\in \llbracket 1, N \rrbracket}\mathscr{L}_{\mathit{CPDCL,i}}^{k}$, where $\mathscr{L}_{\mathit{CPDCL,i}}^{k}$ is

\vspace{-5pt}
\begin{equation}
	\small
	\label{eq:cpdc}
	\begin{aligned}
		&\mathscr{L}_{\mathit{CPDCL,i}}^{(k)}=-\log \frac{\exp \left(\left\langle\vec{\mathit{H}}_{\vec{p}_i}^{(1)}, \vec{\mathit{H}}_{\vec{p}_i}^{(2)}\right\rangle / t\right)}{\sum_{j=1,j \neq i}^{N}\sum_{l=1}^{2}  \exp \left(\left\langle\vec{\mathit{H}}_{\vec{p}_i}^{(k)}, \vec{\mathit{H}}_{\vec{p}_j}^{(l)}\right\rangle / t\right)}	\\
		& =-\frac{\langle \vec{\mathit{H}}_{\vec{p}_i}^{(1)}, \vec{\mathit{H}}_{\vec{p}_i}^{(2)} \rangle}{t}+\log \sum_{j=1,j \neq i}^{N}\sum_{l=1}^{2}  \exp \left(\frac{\left\langle\vec{\mathit{H}}_{\vec{p}_i}^{(k)}, \vec{\mathit{H}}_{\vec{p}_j}^{(l)}\right\rangle}{t}\right)
	\end{aligned}
\end{equation}

Furthermore, we extend the \emph{CPDCL} loss to a weighted variant by introducing a weighting function over the positive pairs. This generalization allows the model to dynamically adjust the contribution of each positive pair during training. The overall weighted loss is defined as:

\begin{equation}
	\begin{aligned}
		\mathscr{L}_{\mathit{CPDCL, i}}^{(k)}=&-w\left(\vec{\mathit{H}}_{\vec{p}_i}^{(1)}, \vec{\mathit{H}}_{\vec{p}_i}^{(2)}\right)\left(\left\langle\vec{\mathit{H}}_{\vec{p}_i}^{(1)}, \vec{\mathit{H}}_{\vec{p}_i}^{(2)}\right\rangle / t\right)+ \\
		&\log \sum_{j=1,j \neq i}^{N}\sum_{l=1}^{2}  \exp \left(\left\langle\vec{\mathit{H}}_{\vec{p}_i}^{(k)}, \vec{\mathit{H}}_{\vec{p}_j}^{(l)}\right\rangle / t\right)
	\end{aligned}
\end{equation}

\noindent
where we intuitively define a weighting function $w(\vec{\mathit{H}}_{\vec{p}_i}^{(1)}, \vec{\mathit{H}}_{\vec{p}_i}^{(2)})=2-\frac{\exp \left(\left\langle\vec{\mathit{H}}_{\vec{p}_i}^{(1)}, \vec{\mathit{H}}_{\vec{p}_i}^{(2)}\right\rangle / \sigma \right)}{E_i \bigl[ \exp \left(\left\langle\vec{\mathit{H}}_{\vec{p}_i}^{(1)}, \vec{\mathit{H}}_{\vec{p}_i}^{(2)}\right\rangle / \sigma \right) \bigr]}$ for the positive pairs in $\mathscr{L}_{\mathit{CPDCL}}$ based on a negative von Mises-Fisher distribution, where $E\big[ w\big] =1$. Intuitively, this weighting emphasizes harder positive pairs---those with lower similarity---by assigning them higher weights, thus providing stronger learning signals. Notably, the original \emph{CPDCL} loss is a special case of the weighted \emph{CPDCL} loss: as $\sigma \to \infty$, the weight approaches 1, and $w\mathscr{L}_{\mathit{CPDCL}} \to \mathscr{L}_{\mathit{CPDCL}}$.

The design of the weighting function is motivated by the imbalance between easy and hard positive pairs in contrastive learning. Positive pairs with high similarity already provide limited optimization signals, while harder positive pairs with lower similarity are more informative for improving representation alignment. The proposed weighting function assigns larger weights to harder positive pairs and smaller weights to easier ones, thereby encouraging the model to focus on under-aligned positive pairs during training. Since the expected weight satisfies $E[w]=1$, the weighted objective preserves the overall loss scale and avoids introducing an additional bias in expectation. Moreover, the original \emph{CPDCL} is recovered when $\sigma \to \infty$, indicating that the weighted variant is a generalized form of \emph{CPDCL}. This design improves training stability while maintaining the decoupling advantage of \emph{CPDCL}.

\subsection{Cross-Node Decoupled Contrastive Learning}
\label{section:local}
In addition to decoupled contrastive learning over global path-level embedding learning, we further incorporate local contrastive signals by introducing a \textit{Cross-Node Decoupled Contrastive loss} (\emph{CNDCL}). This formulation captures fine-grained differences between positive and negative edge-level (node embedding) samples, serving as a strong regularization that improves the embedding capacity of the path encoder. As illustrated in Figure~\ref{fig:overview}, \emph{CNDCL} complements the global cross-path loss by enforcing consistency at a more granular level, thereby enhancing the generalizability of the learned path embeddings.

Following the formulation of the \emph{CPDCL}, the objective of the weighted
\emph{CNDCL} is to maximize the similarity between corresponding edge
embeddings from the two augmented views $\vec{p}_i^{(1)}$ and
$\vec{p}_i^{(2)}$, while minimizing their similarity to negative edge
samples. This encourages local structural alignment across views and improves
the model's ability to capture fine-grained variations. Formally,
\emph{CNDCL} is reformulated as
$\mathscr{L}_{\mathit{CNDCL}}
=\sum_{i=1}^{N}\sum_{h=1}^{|p_i|}\sum_{k=1}^{2}
\mathscr{L}_{\mathit{CNDCL},i,h}^{(k)}$, where $\mathscr{L}_{\mathit{CNDCL}, i, h}^{(k)}$ denotes the decoupled contrastive loss for the $h$-th edge (or node embedding) in the $k$-th view of path $p_i$ and $\mathscr{L}_{\mathit{CNDCL, i, h}}^{(k)}$ is	
\begin{align}
\mathscr{L}_{\mathit{CNDCL},i,h}^{(k)}
&= -\sum_{h=1}^{|p_i|} 
w\!\left(\vec{p}_{i,h}^{(1)}, \vec{p}_{i,h}^{(2)}\right)
\left(\langle \vec{p}_{i,h}^{(1)}, \vec{p}_{i,h}^{(2)} \rangle / t \right) \notag \\
&\quad + \log \sum_{\substack{j=1 \\ j \neq i}}^{N} 
\sum_{h=1}^{|p_i|} \sum_{l=1}^{2}  
\exp \left(\langle \vec{p}_{i,h}^{(k)}, \vec{p}_{j,h}^{(l)} \rangle /t \right)
\tag{11}\label{eq:cndcl}
\end{align}

\section{Adaptive Co-training}
\label{sec:ACT}
To efficiently train both the view generator and path encoder in an end-to-end manner, we jointly optimize four key loss components: the lower bound and reconstruction losses from the view generator ($\mathscr{L}_{\mathit{LB}}$ and $\mathscr{L}_{\mathit{Rec}}$), the global-level contrastive loss ($\mathscr{L}_{\mathit{CPDCL}}$), and the local-level contrastive loss ($\mathscr{L}_{\mathit{CNDCL}}$). A common approach is to combine these objectives through a weighted sum:
\begin{equation}
\mathscr{L}_{\mathit{Total}} =\lambda_1 \mathscr{L}_{\mathit{LB}}+\lambda_2 \mathscr{L}_{\mathit{Rec}}+\lambda_3 \mathscr{L}_{\mathit{CPDCL}} +\lambda_4 \mathscr{L}_{\mathit{CNDCL}}
\end{equation}
\noindent
where $\lambda_1$, $\lambda_2$, $\lambda_3$, and $\lambda_4$ are manually selected hyperparameters. These weights are typically chosen based on empirical tuning or heuristic rules. However, this manual process is both time-consuming and sub-optimal, especially in complex multi-objective settings where the relative importance of each loss can vary dynamically over the course of training. The underlying intuition is that multiple training losses should be adaptively weighted in response to task uncertainty and training dynamics, thereby promoting balanced optimization and enhancing overall performance.

To address this limitation, we introduce an adaptive co-training strategy that learns to weight each loss automatically based on task-specific uncertainty. Inspired by Ref. \cite{DBLP:conf/cvpr/KendallGC18}, we formulate the total loss as:

\vspace{-5pt}
\begin{equation}
\begin{aligned}
\mathscr{L}_{\mathit{Total}} =&\frac{1}{2 \sigma_1^{2}}\mathscr{L}_{\mathit{LB}}+\frac{1}{2 \sigma_2^{2}}\mathscr{L}_{\mathit{Rec}}+\frac{1}{2 \sigma_3^{2}}\mathscr{L}_{\mathit{CPDCL}} + \\ 
&\frac{1}{2 \sigma_4^{2}} \mathscr{L}_{\mathit{CNDCL}}
+(\log \sigma_1 \sigma_2 \sigma_3 \sigma_4)
\end{aligned}
\end{equation}
\noindent
where $\sigma_i$ is a learnable uncertainty parameter for the $i$-th loss component. A larger $\sigma_i$ reduces the weight of the corresponding loss, while the logarithmic regularization term prevents $\sigma_i$ from increasing without bound.

This uncertainty-aware formulation offers two key benefits: (i) dynamic adaptation: as training progresses and the difficulty of each objective changes, the model adjusts its focus accordingly—allocating more learning capacity to losses that offer more stable and informative gradients; (ii) improved stability and performance: by eliminating manual tuning and mitigating loss-scale imbalance, this strategy leads to more stable convergence and better generalization across both global and local levels of path embedding learning.

\section{Complexity Analysis}
In this section, we provide a theoretical complexity analysis of \emph{AutoPRL}. Let $N$ denote the batch size, $M$ the number of nodes in a path, $d_n$ the node embedding dimension, $d_h$ the VNAE hidden dimension, $d_z$ the latent dimension, $d_p$ the path embedding dimension, and $K$ the number of Transformer layers. The complexity of \emph{AutoPRL} mainly comes from the VNAE-based view generator, the Transformer-based path encoder, and the decoupled contrastive objectives.

For the VNAE-based view generator, the LSTM encoder and decoder process node-embedding sequences of length $M$. The encoder has time complexity $O(NM(d_nd_h+d_h^2))$, while the decoder has complexity $O(NM(d_zd_h+d_h^2+d_hd_n))$. The linear layers for estimating the mean, variance, and latent-to-hidden mapping add $O(Nd_hd_z)$ overhead, which is minor compared with the recurrent computation. Thus, the overall time complexity of the VNAE module is $O(NM(d_nd_h+d_zd_h+d_h^2+d_hd_n))$.

For the Transformer-based path encoder, each view is prepended with a CLS token and processed by $K$ Transformer layers. The complexity of one Transformer layer is dominated by self-attention and feed-forward networks, i.e., $O(NM^2d_p+NMd_p^2)$. Since \emph{AutoPRL} encodes two views, the total Transformer encoding complexity is $O(2KN(M^2d_p+Md_p^2))$, which dominates the runtime when $M$ or $K$ is large.

For the decoupled contrastive objectives, the global path-level loss costs $O(N^2d_p)$, while the local node-level loss costs $O(MN^2d_p)$. Therefore, the total contrastive-learning complexity is $O((M+1)N^2d_p)$. The ACT module only introduces three learnable scalar weights and thus has negligible runtime overhead.

In terms of parameter memory complexity, the VNAE module requires $O(d_nd_h+d_zd_h+d_h^2+d_hd_z)$ parameters, and the Transformer path encoder requires $O(Kd_p^2)$ parameters. The contrastive objectives introduce no additional trainable parameters, while ACT adds only three scalars. Therefore, the total parameter memory complexity is $O(d_nd_h+d_zd_h+d_h^2+d_hd_z+Kd_p^2)$. Overall, the computational and memory costs of \emph{AutoPRL} are dominated by the VNAE view generator and Transformer encoder, whereas the additional overhead of \emph{CVDC} and \emph{ACT} is limited.

\section{Experimental Evaluation}
To demonstrate the effectiveness of \emph{AutoPRL}, we conduct experiments on three real-world trajectory datasets.
Our focus is on five key downstream tasks: travel time estimation, path ranking, most similar path search, path recommendation, and path imputation. For each dataset, we compare \emph{AutoPRL}’s performance with state-of-the-art baselines.
\subsection{Experimental Setup}

\subsubsection{Dataset} We evaluate our approach using three real-world datasets: Aalborg (Denmark) \cite{DBLP:journals/tkde/YangGY22}, Chengdu (China) \cite{DBLP:journals/pvldb/00020B22} and Harbin (China) \cite{DBLP:conf/www/LiCSC19}. The Aalborg road network contains 10,017 nodes and 11,597 edges, the Chengdu road network comprises 6,632 nodes and 17,038 edges, and the Harbin road network contains 8,497 nodes and 14,497 edges. We also collect three large-scale GPS trajectory datasets corresponding to these networks. Specifically, we use 33,264 path trajectories in the Aalborg dataset, 54,491 paths in the Chengdu dataset, and 259,615 paths in the Harbin dataset after map matching \cite{DBLP:journals/gis/YangG18,DBLP:conf/icde/YangGHYTJ22,DBLP:conf/icde/LiZCJW18,DBLP:conf/aaai/HanYH26}. We also remove the paths with fewer than 20 nodes.
The underlying road network graphs for three cities are obtained from OpenStreetMap\footnote{\href{https://www.openstreetmap.org}{https://www.openstreetmap.org}}, ensuring real-world topological consistency and scalability for our experiments.

\subsubsection{Downstream Tasks} \textbf{Travel Time Estimation: }For each path, we extract the corresponding travel time (in minutes) from the GPS trajectory data. The goal is to apply a regression model that predicts the travel time based on the learned path embedding vectors. To evaluate the performance of travel time estimation, we adopt three widely used metrics: Mean Absolute Error (MAE), Mean Absolute Percentage Error (MAPE), and Root Mean Square Error (RMSE).
\noindent
\textbf{Path Ranking: }Given a set of candidate paths—typically sharing the same source and destination—each path is assigned a ranking score within the range [0, 1], reflecting its relative preference based on historical trajectory data. Following prior studies \cite{DBLP:journals/tkde/YangGY22,DBLP:conf/kdd/YangHGYJ23}, we compute these scores using similarity measurements derived from real-world trajectories. To evaluate the performance of path ranking, we adopt three widely used metrics: Mean Absolute Error (MAE), the Kendall rank correlation coefficient ($\tau$), and the Spearman rank correlation coefficient ($\rho$). These metrics collectively assess both the accuracy of the predicted ranking scores and the consistency of the predicted orderings relative to the ground truth.

\noindent
\textbf{Most Similar Path Search: }
Given a query path and a path database, the goal is to retrieve the path that is most similar to the query based on the learned path embeddings. We use Mean Rank as the evaluation metric, where a lower value indicates better retrieval performance.

\noindent
\textbf{Path Recommendation: }
Given a set of candidate paths, the goal is to identify the path preferred by users based on historical trajectory data. We formulate this task as a binary classification problem, where the actually traveled path is regarded as a positive sample and alternative paths are regarded as negative samples. We evaluate the performance using Accuracy, Precision, and F1 score.

\noindent
\textbf{Path Imputation: }
Given an incomplete path with a portion of its segments masked, the goal is to recover the missing path segments using the learned path representations. We randomly mask 30\% of the path segments and use the original path as the ground truth. We adopt Recall@3 (R@3), Recall@5 (R@5), and Mean Average Precision (MAP) as the evaluation metrics.

\subsubsection{Baselines} We evaluate \emph{AutoPRL} with 13 baselines, consisting of 9 unsupervised learning-based methods and 4 supervised learning-based methods. The unsupervised learning based methods are: \emph{Node2vec}~\cite{DBLP:conf/kdd/GroverL16}, 
\emph{VGAE-AMI}~\cite{DBLP:journals/tkdd/LiLL23},
\emph{DVGA}~\cite{DBLP:journals/corr/abs-2402-01143},
\emph{Toast}~\cite{DBLP:conf/cikm/ChenLCBLLCE21}, \emph{PIM}~\cite{DBLP:conf/ijcai/YangGHT021}, \emph{START}~\cite{DBLP:conf/icde/JiangPRJLW23}, \emph{LightPath}~\cite{DBLP:conf/kdd/YangHGYJ23}, 
\emph{TrajRL}~\cite{DBLP:conf/icde/XiaZCCYD25},
and \emph{RED}~\cite{DBLP:journals/pvldb/ZhouSCJK24}.
The supervised methods take into account the labels from a specific downstream task to
obtain path representations, which are: \emph{PathRank}~\cite{DBLP:journals/tkde/YangGY22}, \emph{HierETA}~\cite{DBLP:conf/kdd/ChenXGFMCC22}, \emph{CompactETA}~\cite{DBLP:conf/kdd/FuMYW20}, and \emph{HMTRL}~\cite{DBLP:journals/vldb/LiuHFLCX23}. 

For all unsupervised learning methods, we first train the path encoder using unlabeled trajectory data—specifically, 30K unlabeled samples from the Aalborg dataset, 50K from the Chengdu dataset and 200K from the Harbin dataset—to learn path embedding vectors. These embeddings are then used as input to a regression model trained on a smaller, labeled dataset (e.g., 12K labeled paths from Aalborg) to predict travel time and path ranking scores. For the regression model, we adopt the ensemble-based Gradient Boosting Regressor (GBR)~\cite{DBLP:conf/nips/PeterDHN17} as the prediction model since our tasks are regression tasks.
In contrast, for all supervised learning methods, the path encoder is trained directly using the limited labeled training data (e.g., the 12K labeled samples from the Aalborg dataset), without access to the larger unlabeled set.

\begin{table*}[t]

\caption{Overall performance on travel time estimation. The best results are highlighted in bold, and the second-best results are underlined.}
\centering
{\renewcommand{\arraystretch}{1.3}
\begin{tabular}{ll|lll|lll|lll}
	\toprule[1.5pt]
	\multicolumn{2}{c|}{\multirow{2}{*}{\textbf{Method}}}                             & \multicolumn{3}{c|}{\textbf{Aalborg}}                                                  & \multicolumn{3}{c|}{\textbf{Chengdu}}                                                  & \multicolumn{3}{c}{\textbf{Harbin}}                                                   \\ \cline{3-11} 
	\multicolumn{2}{l|}{}                                                    & \multicolumn{1}{l}{\textbf{MAE}$\downarrow$} & \multicolumn{1}{l}{\textbf{MAPE}(\%)$\downarrow$} & \multicolumn{1}{l|}{\textbf{RMSE}$\downarrow$} & \multicolumn{1}{l}{\textbf{MAE}$\downarrow$} & \multicolumn{1}{l}{\textbf{MAPE}(\%)$\downarrow$} & \multicolumn{1}{l|}{\textbf{RMSE}$\downarrow$} & \multicolumn{1}{l}{\textbf{MAE}$\downarrow$} & \multicolumn{1}{l}{\textbf{MAPE}(\%)$\downarrow$} &  \multicolumn{1}{l}{\textbf{RMSE}$\downarrow$ }\\ \toprule[0.9pt]
	\multicolumn{1}{l|}{\multirow{4}{*}{\rotatebox[origin=c]{90}{Supervised}}}   
	& \textit{PathRank}      & \multicolumn{1}{c}{2.913}             & \multicolumn{1}{c}{18.713}              &  \multicolumn{1}{c|}{4.832}              & \multicolumn{1}{c}{5.802}             & \multicolumn{1}{c}{34.014}              & \multicolumn{1}{c|}{8.894}               & \multicolumn{1}{c}{4.127}             & \multicolumn{1}{c}{14.339}              & \multicolumn{1}{c}{5.301}               \\ \cline{2-11} 
	
	\multicolumn{1}{l|}{}                              & \textit{HierETA} & \multicolumn{1}{c}{3.097}             & \multicolumn{1}{c}{18.620}              & \multicolumn{1}{c|}{5.058}            & \multicolumn{1}{c}{5.060}             & \multicolumn{1}{c}{28.820}              & \multicolumn{1}{c|}{7.921}               & \multicolumn{1}{c}{4.897}             & \multicolumn{1}{c}{16.506}              &\multicolumn{1}{c}{5.961}                \\ \cline{2-11} 

	\multicolumn{1}{l|}{}                              & \textit{CompactETA} & \multicolumn{1}{c}{3.030}             & \multicolumn{1}{c}{19.726}              & \multicolumn{1}{c|}{4.791}              & \multicolumn{1}{c}{\underline{4.832}}             & \multicolumn{1}{c}{28.678}              & \multicolumn{1}{c|}{7.447}              & \multicolumn{1}{c}{4.372}             & \multicolumn{1}{c}{14.974}              &  \multicolumn{1}{c}{5.362}              \\ \cline{2-11} 
	
	\multicolumn{1}{l|}{}                              & \textit{HMTRL}    & \multicolumn{1}{c}{3.515}             & \multicolumn{1}{c}{19.955}              &\multicolumn{1}{c|}{5.792 }              & \multicolumn{1}{c}{4.861}             & \multicolumn{1}{c}{28.529}              &\multicolumn{1}{c|}{7.473}               & \multicolumn{1}{c}{4.704}             & \multicolumn{1}{c}{15.543}              & \multicolumn{1}{c}{5.803}                     \\ \hline
	
	\multicolumn{1}{l|}{\multirow{9}{*}{\rotatebox[origin=c]{90}{Unsupervised}}} & \textit{Node2vec}   & \multicolumn{1}{c}{5.701}       & \multicolumn{1}{c}{35.485}          &  \multicolumn{1}{c|}{8.165}         & \multicolumn{1}{c}{5.997}       & \multicolumn{1}{c}{35.919}          &  \multicolumn{1}{c|}{8.653}         & \multicolumn{1}{c}{5.399}       & \multicolumn{1}{c}{18.709}          & \multicolumn{1}{c}{6.773}         \\ \cline{2-11} 
	
	\multicolumn{1}{l|}{}                              & \textit{VGAE-AMI}      & \multicolumn{1}{c}{4.986}             & \multicolumn{1}{c}{32.421}              &\multicolumn{1}{c|}{7.458}               & \multicolumn{1}{c}{5.454}             & \multicolumn{1}{c}{34.192}              &\multicolumn{1}{c|}{7.860}               & \multicolumn{1}{c}{5.193}             & \multicolumn{1}{c}{17.824}              &\multicolumn{1}{c}{6.359}               \\ \cline{2-11} 
	
	\multicolumn{1}{l|}{}                              & \textit{DVGA}      & \multicolumn{1}{c}{5.022}             & \multicolumn{1}{c}{33.128}              &\multicolumn{1}{c|}{7.846}               & \multicolumn{1}{c}{5.601}             & \multicolumn{1}{c}{34.892}              &\multicolumn{1}{c|}{8.011}               & \multicolumn{1}{c}{5.241}             & \multicolumn{1}{c}{18.198}              &\multicolumn{1}{c}{6.514}               \\ \cline{2-11} 
	
	\multicolumn{1}{l|}{}                              & \textit{Toast}      & \multicolumn{1}{c}{3.439}             & \multicolumn{1}{c}{21.311}              &\multicolumn{1}{c|}{5.555}               & \multicolumn{1}{c}{4.918}             & \multicolumn{1}{c}{28.567}              &\multicolumn{1}{c|}{7.513}               & \multicolumn{1}{c}{4.627}             & \multicolumn{1}{c}{16.101}              &\multicolumn{1}{c}{5.803}               \\ \cline{2-11} 
	\multicolumn{1}{l|}{}                              & \textit{PIM}        & \multicolumn{1}{c}{3.471}             & \multicolumn{1}{c}{21.314}              &\multicolumn{1}{c|}{5.640}               & \multicolumn{1}{c}{4.924}             & \multicolumn{1}{c}{28.553}              &\multicolumn{1}{c|}{7.493}               & \multicolumn{1}{c}{4.538}             & \multicolumn{1}{c}{15.713}              &\multicolumn{1}{c}{5.682}               \\ \cline{2-11} 
	
	\multicolumn{1}{l|}{}                              & \textit{START}      & \multicolumn{1}{c}{2.965}             & \multicolumn{1}{c}{18.432}              &\multicolumn{1}{c|}{4.979}               & \multicolumn{1}{c}{5.174}             & \multicolumn{1}{c}{30.769}              &\multicolumn{1}{c|}{7.726}                 & \multicolumn{1}{c}{4.402}             & \multicolumn{1}{c}{15.262}              &\multicolumn{1}{c}{5.503}                 \\ \cline{2-11} 
	
	\multicolumn{1}{l|}{}                              & \textit{LightPath}  & \multicolumn{1}{c}{2.920}             & \multicolumn{1}{c}{18.254}              &\multicolumn{1}{c|}{4.932}               & \multicolumn{1}{c}{4.850}             & \multicolumn{1}{c}{28.387}              &\multicolumn{1}{c|}{7.406}               & \multicolumn{1}{c}{4.221}             & \multicolumn{1}{c}{14.962}              &\multicolumn{1}{c}{5.310}               \\ \cline{2-11}

	\multicolumn{1}{l|}{}                              & \textit{RED}       & \multicolumn{1}{c}{2.457}             & \multicolumn{1}{c}{14.907}             &\multicolumn{1}{c|}{3.809}               & \multicolumn{1}{c}{4.845}             & \multicolumn{1}{c}{28.380}             &\multicolumn{1}{c|}{7.296}                 & \multicolumn{1}{c}{4.097}           & \multicolumn{1}{c}{14.214}             &\multicolumn{1}{c}{5.153}              \\ \cline{2-11} 
		
	\multicolumn{1}{l|}{}                              & \textit{TrajRL}       & \multicolumn{1}{c}{\underline{2.243}}             & \multicolumn{1}{c}{\underline{13.125}}              &\multicolumn{1}{c|}{\underline{3.632}}               & \multicolumn{1}{c}{\underline{4.724}}             & \multicolumn{1}{c}{\underline{27.960}}              &\multicolumn{1}{c|}{\underline{7.108}}                 & \multicolumn{1}{c}{\underline{3.994}}             & \multicolumn{1}{c}{\underline{13.872}}              &\multicolumn{1}{c}{\underline{5.107}}                 \\ \cline{2-11} 
		
	\multicolumn{1}{l|}{}                              & \textit{AutoPRL}      & \multicolumn{1}{c}{\textbf{1.971}}             & \multicolumn{1}{c}{\textbf{12.867}}              &\multicolumn{1}{c|}{\textbf{3.382}}               & \multicolumn{1}{c}{\textbf{4.672}}             & \multicolumn{1}{c}{\textbf{27.318}}              &\multicolumn{1}{c|}{\textbf{7.074}}                 & \multicolumn{1}{c}{\textbf{3.883}}             & \multicolumn{1}{c}{\textbf{13.421}}              &\multicolumn{1}{c}{\textbf{4.896}}                 \\ \hline
		\multicolumn{1}{l|}{}                              & \textit{Imp. }           & \multicolumn{1}{c}{12.13\%}             & \multicolumn{1}{c}{1.97\%}              &\multicolumn{1}{c|}{6.88\%}               & \multicolumn{1}{c}{1.10\%}             & \multicolumn{1}{c}{2.30\%}              &\multicolumn{1}{c|}{0.48\%}                 & \multicolumn{1}{c}{2.78\%}             & \multicolumn{1}{c}{3.25\%}              &\multicolumn{1}{c}{4.13\%} \\ \bottomrule[1.5pt]
	\end{tabular}}
	\label{tab:miantte}
\end{table*}

\begin{table*}[htp]
\caption{Overall performance on path ranking. The best results are highlighted in bold, and the second-best results are underlined.}
\centering
{\renewcommand{\arraystretch}{1.3}
\begin{tabular}{ll|lll|lll|lll}
	\toprule[1.5pt]
	\multicolumn{2}{c|}{\multirow{2}{*}{\textbf{Method}}}                             & \multicolumn{3}{c|}{\textbf{Aalborg}}                                                  & \multicolumn{3}{c|}{\textbf{Chengdu}}                                                  & \multicolumn{3}{c}{\textbf{Harbin}}                                                   \\ \cline{3-11} 
	\multicolumn{2}{l|}{}                                                    & \multicolumn{1}{l}{\textbf{MAE}$\downarrow$} & \multicolumn{1}{l}{\textbf{TAU}($\tau$)$\uparrow$} & \textbf{RHO}($\rho$)$\uparrow$ & \multicolumn{1}{l}{\textbf{MAE}$\downarrow$} & \multicolumn{1}{l}{\textbf{TAU}($\tau$)$\uparrow$} & \textbf{RHO}($\rho$)$\uparrow$ & \multicolumn{1}{l}{\textbf{MAE}$\downarrow$} & \multicolumn{1}{l}{\textbf{TAU}($\tau$)$\uparrow$} & \textbf{RHO}($\rho$)$\uparrow$ \\ \toprule[0.9pt]
	\multicolumn{1}{l|}{\multirow{4}{*}{\rotatebox[origin=c]{90}{Supervised}}}    & \textit{PathRank}   & \multicolumn{1}{c}{0.159}             & \multicolumn{1}{c}{0.727}              &\multicolumn{1}{c|}{0.751}               & \multicolumn{1}{c}{0.123}            & \multicolumn{1}{c}{0.703}              &\multicolumn{1}{c|}{0.742}                 & \multicolumn{1}{c}{0.145}             & \multicolumn{1}{c}{0.821}              &\multicolumn{1}{c}{0.832}                 \\ \cline{2-11} 
	
	\multicolumn{1}{l|}{}                               & \textit{HierETA}   & \multicolumn{1}{c}{0.152}             & \multicolumn{1}{c}{0.757}              &\multicolumn{1}{c|}{0.767}               & \multicolumn{1}{c}{0.178}             & \multicolumn{1}{c}{0.735}              &\multicolumn{1}{c|}{0.774}                 & \multicolumn{1}{c}{0.133}             & \multicolumn{1}{c}{0.846}              &\multicolumn{1}{c}{0.879}               \\ \cline{2-11} 
	
	\multicolumn{1}{l|}{}                               & \textit{CompactETA}   & \multicolumn{1}{c}{0.169}             & \multicolumn{1}{c}{0.698}              &\multicolumn{1}{c|}{0.709}               & \multicolumn{1}{c}{0.195}             & \multicolumn{1}{c}{0.661}              &\multicolumn{1}{c|}{0.705}                 & \multicolumn{1}{c}{0.160}             & \multicolumn{1}{c}{0.796}              &\multicolumn{1}{c}{0.833}               \\ \cline{2-11} 
	
	\multicolumn{1}{l|}{}                               & \textit{HMTRL}   & \multicolumn{1}{c}{0.163}             & \multicolumn{1}{c}{0.729}              &\multicolumn{1}{c|}{0.741}               & \multicolumn{1}{c}{0.149}             & \multicolumn{1}{c}{0.779}              &\multicolumn{1}{c|}{0.789}                 & \multicolumn{1}{c}{0.133}             & \multicolumn{1}{c}{0.838}              &\multicolumn{1}{c}{0.874}                        \\ \hline

	\multicolumn{1}{l|}{\multirow{9}{*}{\rotatebox[origin=c]{90}{Unsupervised}}}  & \textit{Node2vec}      & \multicolumn{1}{c}{0.253}             & \multicolumn{1}{c}{0.632}              &\multicolumn{1}{c|}{0.637}               & \multicolumn{1}{c}{0.246}             & \multicolumn{1}{c}{0.464}              &\multicolumn{1}{c|}{0.505}                 & \multicolumn{1}{c}{0.211}             & \multicolumn{1}{c}{0.468}              &\multicolumn{1}{c}{0.512}                  \\ \cline{2-11}  
	
		\multicolumn{1}{l|}{}                              & \textit{VGAE-AMI}      & \multicolumn{1}{c}{0.194}             & \multicolumn{1}{c}{0.662}              &\multicolumn{1}{c|}{0.685}               & \multicolumn{1}{c}{0.212}             & \multicolumn{1}{c}{0.543}              &\multicolumn{1}{c|}{0.577}               & \multicolumn{1}{c}{0.188}             & \multicolumn{1}{c}{0.494}              &\multicolumn{1}{c}{0.524}               \\ \cline{2-11} 
	
	\multicolumn{1}{l|}{}                              & \textit{DVGA}      & \multicolumn{1}{c}{0.211}             & \multicolumn{1}{c}{0.653}              &\multicolumn{1}{c|}{0.669}               & \multicolumn{1}{c}{0.220}             & \multicolumn{1}{c}{0.522}              &\multicolumn{1}{c|}{0.549}               & \multicolumn{1}{c}{0.195}             & \multicolumn{1}{c}{0.475}              &\multicolumn{1}{c}{0.501}               \\ \cline{2-11} 
	
	\multicolumn{1}{l|}{}                              & \textit{Toast}      & \multicolumn{1}{c}{0.138}             & \multicolumn{1}{c}{0.768}              &\multicolumn{1}{c|}{0.779}               & \multicolumn{1}{c}{0.131}             & \multicolumn{1}{c}{0.734}              &\multicolumn{1}{c|}{0.752}                 & \multicolumn{1}{c}{0.161}             & \multicolumn{1}{c}{0.840}              &\multicolumn{1}{c}{0.878}                \\ \cline{2-11} 
	
	\multicolumn{1}{l|}{}                              & \textit{PIM}      & \multicolumn{1}{c}{0.141}             & \multicolumn{1}{c}{0.764}              &\multicolumn{1}{c|}{0.777}               & \multicolumn{1}{c}{0.125}             & \multicolumn{1}{c}{0.725}              &\multicolumn{1}{c|}{0.745}                 & \multicolumn{1}{c}{0.119}             & \multicolumn{1}{c}{0.879}              &\multicolumn{1}{c}{0.911}                 \\ \cline{2-11} 
	
	\multicolumn{1}{l|}{}                              & \textit{START}      & \multicolumn{1}{c}{0.178}             & \multicolumn{1}{c}{0.839}              &\multicolumn{1}{c|}{0.859}              & \multicolumn{1}{c}{0.228}             & \multicolumn{1}{c}{0.638}              &\multicolumn{1}{c|}{0.679}                 & \multicolumn{1}{c}{0.151}             & \multicolumn{1}{c}{0.859}              &\multicolumn{1}{c}{0.890}                 \\ \cline{2-11} 
	
	\multicolumn{1}{l|}{}                              & \textit{LightPath}  & \multicolumn{1}{c}{0.118}             & \multicolumn{1}{c}{0.781}              &\multicolumn{1}{c|}{0.791}               & \multicolumn{1}{c}{0.195}             & \multicolumn{1}{c}{0.711}              &\multicolumn{1}{c|}{0.748}                 & \multicolumn{1}{c}{0.122}             & \multicolumn{1}{c}{0.881}             &\multicolumn{1}{c}{0.913}                 \\ \cline{2-11} 
	
	\multicolumn{1}{l|}{}                              & \textit{RED}        & \multicolumn{1}{c}{0.109}             & \multicolumn{1}{c}{0.819}              &\multicolumn{1}{c|}{0.823}               & \multicolumn{1}{c}{0.129}             & \multicolumn{1}{c}{0.797}           &\multicolumn{1}{c|}{0.833}                 & \multicolumn{1}{c}{0.127}             & \multicolumn{1}{c}{0.858}              &\multicolumn{1}{c}{0.865}                 \\ \cline{2-11} 
	
	\multicolumn{1}{l|}{}                              & \textit{TrajRL}        & \multicolumn{1}{c}{\underline{0.089}}             & \multicolumn{1}{c}{\underline{0.855}}              &\multicolumn{1}{c|}{\underline{0.871}}               & \multicolumn{1}{c}{\underline{0.114}}             & \multicolumn{1}{c}{\underline{0.810}}              &\multicolumn{1}{c|}{\underline{0.839}}                 & \multicolumn{1}{c}{\underline{0.116}}             & \multicolumn{1}{c}{\underline{0.896}}              &\multicolumn{1}{c}{\underline{0.929}}                 \\ \cline{2-11} 
	
	\multicolumn{1}{l|}{}                             & \textit{AutoPRL}            & \multicolumn{1}{c}{\textbf{0.073}}             & \multicolumn{1}{c}{\textbf{0.890}}              &\multicolumn{1}{c|}{\textbf{0.893}}               & \multicolumn{1}{c}{\textbf{0.103}}             & \multicolumn{1}{c}{\textbf{0.836}}              &\multicolumn{1}{c|}{\textbf{0.845} }                & \multicolumn{1}{c}{\textbf{0.105}}             & \multicolumn{1}{c}{\textbf{0.939}}              &\multicolumn{1}{c}{\textbf{0.958}}                  \\ \hline
	
	\multicolumn{1}{l|}{}                              & \textit{Imp.}       & \multicolumn{1}{c}{17.98\%}             & \multicolumn{1}{c}{4.09\%}              &\multicolumn{1}{c|}{2.53\%}               & \multicolumn{1}{c}{9.65\%}             & \multicolumn{1}{c}{3.21\%}              &\multicolumn{1}{c|}{0.72\%}                 & \multicolumn{1}{c}{9.48\%}             & \multicolumn{1}{c}{4.80\%}              &\multicolumn{1}{c}{3.12\%}                 \\ \bottomrule[1.5pt]
	\end{tabular}}
	\label{tab:mianpr}
	\vspace{-10pt} 
\end{table*}
\subsubsection{Implementation Details} We implement \emph{AutoPRL} using PyTorch 1.12.0. We use a machine running a Linux server with 40 Intel(R) Xeon(R) W-2155 CPU @ 3.30GHz and four Quadro RTX 8000 GPU cards. We set the embedding size $d_p$ to 128 in all methods. 
We optimize the learnable view generator and path encoder using the AdamW optimizer, paired with a cosine decay learning rate schedule. The training batch size is set to 64.
The base learning rate is set to 1e-5 for the Aalborg and Chengdu datasets, and 1e-6 for the Harbin dataset. The source code is available at \url{https://github.com/Sean-Bin-Yang/Path-Representation-Learning-Library/tree/master/AutoPRL}. 

\subsection{Main Results}
The main experimental results are reported in Table~\ref{tab:miantte} for travel time estimation, Table~\ref{tab:mianpr} for path ranking, Table~\ref{tab:SC} for most similar path search, Table~\ref{tab:PRP} for path recommendation, and Table~\ref{tab:NLP} for path imputation.

\noindent
\textbf{Performance at Travel Time Estimation. }Table \ref{tab:miantte} reports the overall performance of our \emph{AutoPRL} and the baseline methods across the three real-world datasets, evaluated using various metrics. Specifically, we select 30K unlabeled paths from the Aalborg dataset, 50K from Chengdu (CD), and 200K from Harbin. However, only 12K labeled paths are available for each dataset.
Thus, we use 30K, 50K, and 200K unlabeled paths from the Aalborg, Chengdu (CD), and Harbin datasets, respectively, to train the path encoder for unsupervised methods. In contrast, supervised approaches can only utilize the 12K labeled paths for each dataset.
Overall, Table~\ref{tab:miantte} shows that \emph{AutoPRL} achieves the best observed performance across all metrics on the three real-world datasets. The improvement is most pronounced on Aalborg, followed by Harbin, while the gains on Chengdu are relatively modest. In particular, compared with the strongest baseline, the improvements in MAE and RMSE on Chengdu are 1.10\% and 0.48\%, respectively.

The relatively larger improvement on Aalborg may be related to its dataset
characteristics. Specifically, Aalborg contains the largest road network
among the three datasets, with 10,017 nodes, while only 30K unlabeled paths
are used for representation learning. In comparison, Chengdu and Harbin use
50K and 200K unlabeled paths over road networks with 6,632 and 8,497 nodes,
respectively. This suggests relatively sparser trajectory coverage with
respect to the road-network size on Aalborg. Under such a setting,
pre-defined node masking or dropping may provide less effective augmentation,
whereas \emph{AutoDA} learns feature-level variations directly from
topology-aware node embeddings without explicitly removing path nodes.
This difference may leave greater room for representation improvement on
Aalborg. We regard this as a dataset-specific empirical observation rather than a universal advantage of \emph{AutoPRL}.

Specifically, we observe the following:
First, the graph representation learning-based approach \emph{Node2vec} is less effective in capturing spatial correlations in paths. In contrast, \emph{AutoPRL} effectively models spatial dependencies through its self-attention mechanism, resulting in better empirical performance.
Second, \emph{AutoPRL} achieves better empirical performance than generative self-supervised learning methods such as \emph{Toast} and \emph{RED}, which mainly consider local trajectory information during path encoder training. In comparison, \emph{AutoPRL} integrates both global (see Section~\ref{section:global}) and local (see Section~\ref{section:local}) information.
\emph{AutoPRL} also achieves better overall empirical performance than contrastive learning-based methods such as \emph{PIM}, \emph{START}, and \emph{LightPath}.
In addition, all these unsupervised baselines optimize multiple losses using fixed weights, whereas we propose an adaptive co-training module that dynamically optimizes multiple losses.
Finally, \emph{PathRank}, \emph{HierETA}, \emph{CompactETA}, and \emph{HMTRL} are four supervised learning methods that exhibit relatively lower performance due to the limited availability of labeled training data, as labeling is both time-consuming and costly. This paper considers a scenario where labeled data is scarce.

\noindent
\textbf{Performance at Path Ranking. }
The results for path ranking are presented in Table \ref{tab:mianpr}. \emph{AutoPRL} achieves the best performance across all evaluation metrics, underscoring the effectiveness of its integrated advanced technologies. Specifically, \emph{AutoPRL} incorporates three key components—\emph{AutoDA}, cross-view decoupled contrastive learning, and the Adaptive Co-Training module—which work in concert to enhance generalization of the learned path embeddings.
In particular, the \emph{AutoDA} module in \emph{AutoPRL} automatically generates diverse path views based on a Gaussian distribution. This data-driven augmentation strategy enables the model to explore a richer variety of views during training, improving generality and reducing reliance on predefined augmentation schemes, which are often suboptimal and inflexible.
Moreover, the cross-view decoupled contrastive learning mechanism facilitates a more comprehensive understanding of path representations by decoupling information across views. This allows \emph{AutoPRL} to effectively capture both global and local dependencies, akin to its performance gains in travel time estimation, while avoiding the limitations of traditional contrastive learning methods that rely solely on static augmentation strategies.
Additionally, the adaptive co-training module dynamically adjusts the weighting of multiple loss components during training, effectively balancing them based on the learning dynamics. 

\begin{figure*}[t]
\centering
\includegraphics[scale=0.32]{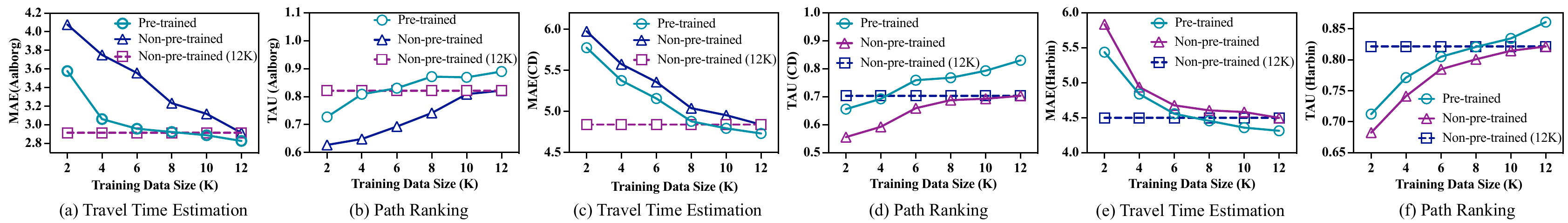}
\caption{Effect of Pre-training.}
\label{fig:pre-train}
\end{figure*}

\begin{table}[t]
	\centering
	\caption{Evaluation on Most Similar Path Search Task.}
	{\renewcommand{\arraystretch}{1.3}
		\begin{tabular}{l|lll|lll}
			\toprule[1.5pt]
			\multirow{2}{*}{\textbf{Method}} & \multicolumn{3}{l|}{\textbf{Chengdu}}                                & \multicolumn{3}{l}{\textbf{Harbin}}                                \\ \cline{2-7} 
			& \multicolumn{1}{l|}{\textbf{10K}} & \multicolumn{1}{l|}{\textbf{30K}} & \textbf{50K} & \multicolumn{1}{l|}{\textbf{60K}} & \multicolumn{1}{l|}{\textbf{80K}} & \textbf{100K}\\ \toprule[1pt]
			\emph{Toast}                   & \multicolumn{1}{l|}{5.233}    & \multicolumn{1}{l|}{7.254}     &9.283      & \multicolumn{1}{l|}{7.674}    & \multicolumn{1}{l|}{9.231}     &11.643      \\ \hline
			\emph{PIM}                     & \multicolumn{1}{l|}{4.107}    & \multicolumn{1}{l|}{6.788}     &8.328      & \multicolumn{1}{l|}{5.962}    & \multicolumn{1}{l|}{7.479}     &9.896      \\ \hline
			\emph{START}                   & \multicolumn{1}{l|}{2.765}    & \multicolumn{1}{l|}{4.371}     & 5.287     & \multicolumn{1}{l|}{4.102}    & \multicolumn{1}{l|}{5.997}     & 8.122     \\ \hline
			\emph{LightPath}               & \multicolumn{1}{l|}{2.431}    & \multicolumn{1}{l|}{3.289}     &5.364      & \multicolumn{1}{l|}{3.562}    & \multicolumn{1}{l|}{5.296}     &7.432      \\ \hline
			\emph{RED}                    & \multicolumn{1}{l|}{1.453}    & \multicolumn{1}{l|}{1.578}     &1.690      & \multicolumn{1}{l|}{2.590}    & \multicolumn{1}{l|}{3.102}     &3.457      \\ \hline
			\emph{TrajRL}                    & \multicolumn{1}{l|}{1.326}    & \multicolumn{1}{l|}{1.467}     &1.582      & \multicolumn{1}{l|}{2.601}    & \multicolumn{1}{l|}{3.229}     &3.757      \\ \hline
			\emph{AutoPRL}                 & \multicolumn{1}{l|}{\textbf{1.217}}    & \multicolumn{1}{l|}{\textbf{1.307}}     &\textbf{1.389}      & \multicolumn{1}{l|}{\textbf{2.571}}    & \multicolumn{1}{l|}{\textbf{2.991}}     &\textbf{3.212}      \\ \toprule[1.5pt]
	\end{tabular}}
	\label{tab:SC}
	\vspace{-10pt}
\end{table}
\noindent

\noindent
\textbf{Performance at Most Similar Path Search. }
The results for most similar path search are reported in Table~\ref{tab:SC}. \emph{AutoPRL} achieves the lowest Mean Rank on both the Chengdu and Harbin datasets across all database sizes, showing consistent performance as the search space increases.
These results indicate that the path representations learned by \emph{AutoPRL} effectively preserve path-level semantic similarity. By capturing both global path semantics and local structural information, \emph{AutoPRL} learns discriminative representations that are effective for similarity-based path retrieval.

\noindent
\textbf{Performance at Path Recommendation. }
The results for path recommendation are presented in Table~\ref{tab:PRP}. \emph{AutoPRL} achieves the best observed performance across Accuracy, Precision, and F1 score on both the Aalborg and Chengdu datasets.
These results indicate that the learned path representations can be effectively applied to the classification-based path recommendation task. By capturing informative path semantics, \emph{AutoPRL} provides discriminative representations for distinguishing user-preferred paths from alternative candidates.

\begin{table}[t]
	\centering
	\caption{Evaluation on Path Recommendation Task.}
	{\renewcommand{\arraystretch}{1.3}
		\begin{tabular}{l|lll|lll}
			\toprule[1.5pt]
			\multirow{2}{*}{\textbf{Method}} & \multicolumn{3}{l|}{\textbf{Aalborg}}                                & \multicolumn{3}{l}{\textbf{Chengdu}}                                \\ \cline{2-7} 
			& \multicolumn{1}{l|}{\textbf{Acc.}} & \multicolumn{1}{l|}{\textbf{Pre.}} & \textbf{F1} & \multicolumn{1}{l|}{\textbf{Acc.}} & \multicolumn{1}{l|}{\textbf{Pre.}} & \textbf{F1}\\ \toprule[1pt]
			\emph{Toast}                   & \multicolumn{1}{l|}{0.692}    & \multicolumn{1}{l|}{0.711}     &0.738      & \multicolumn{1}{l|}{0.774}    & \multicolumn{1}{l|}{0.785}     &0.806      \\ \hline
			\emph{PIM}                     & \multicolumn{1}{l|}{0.722}    & \multicolumn{1}{l|}{0.739}     &0.762      & \multicolumn{1}{l|}{0.785}    & \multicolumn{1}{l|}{0.799}     &0.813      \\ \hline
			\emph{START}                   & \multicolumn{1}{l|}{0.751}    & \multicolumn{1}{l|}{0.770}     &0.787      & \multicolumn{1}{l|}{0.804}    & \multicolumn{1}{l|}{0.826}     &0.853      \\ \hline
			\emph{LightPath}               & \multicolumn{1}{l|}{0.752}    & \multicolumn{1}{l|}{0.768}     &0.790      & \multicolumn{1}{l|}{0.817}    & \multicolumn{1}{l|}{0.832}     &0.849     \\ \hline
			\emph{RED}                    & \multicolumn{1}{l|}{0.785}    & \multicolumn{1}{l|}{0.811}     &0.830      & \multicolumn{1}{l|}{0.844}    & \multicolumn{1}{l|}{0.860}     & 0.879     \\ \hline
			\emph{TrajRL}                    & \multicolumn{1}{l|}{0.789}    & \multicolumn{1}{l|}{0.820}     &0.835      & \multicolumn{1}{l|}{0.850}    & \multicolumn{1}{l|}{0.862}     & 0.880     \\ \hline
			\emph{AutoPRL}                 & \multicolumn{1}{l|}{\textbf{0.791}}    & \multicolumn{1}{l|}{\textbf{0.823}}     &\textbf{0.841}      & \multicolumn{1}{l|}{\textbf{0.858}}    & \multicolumn{1}{l|}{\textbf{0.869}}     &\textbf{0.881}      \\ \toprule[1.5pt]
	\end{tabular}}
	\label{tab:PRP}
\end{table}

\begin{table}[htp]
	\centering
	\caption{Evaluation on Path Imputation Task.}
	{\renewcommand{\arraystretch}{1.3}
		\begin{tabular}{l|lll|lll}
			\toprule[1.5pt]
			\multirow{2}{*}{\textbf{Method}} & \multicolumn{3}{l|}{\textbf{Aalborg}}                                & \multicolumn{3}{l}{\textbf{Chengdu}}                                \\ \cline{2-7} 
			& \multicolumn{1}{l|}{\textbf{R@3}} & \multicolumn{1}{l|}{\textbf{R@5}} & \textbf{MAP} & \multicolumn{1}{l|}{\textbf{R@3}} & \multicolumn{1}{l|}{\textbf{R@5}} & \textbf{MAP}\\ \toprule[1pt]
			\emph{Toast}                   & \multicolumn{1}{l|}{0.656}    & \multicolumn{1}{l|}{0.705}     &0.697      & \multicolumn{1}{l|}{0.770}    & \multicolumn{1}{l|}{0.797}     &0.800      \\ \hline
			\emph{PIM}                     & \multicolumn{1}{l|}{0.679}    & \multicolumn{1}{l|}{0.716}     &0.663      & \multicolumn{1}{l|}{0.792}    & \multicolumn{1}{l|}{0.837}     &0.845      \\ \hline
			\emph{START}                   & \multicolumn{1}{l|}{0.708}    & \multicolumn{1}{l|}{0.725}     &0.692      & \multicolumn{1}{l|}{0.834}    & \multicolumn{1}{l|}{0.860}     &0.863      \\ \hline
			\emph{LightPath}               & \multicolumn{1}{l|}{0.714}    & \multicolumn{1}{l|}{0.756}     &0.736      & \multicolumn{1}{l|}{0.830}    & \multicolumn{1}{l|}{0.855}     &0.823      \\ \hline
			\emph{RED}                    & \multicolumn{1}{l|}{0.733}    & \multicolumn{1}{l|}{0.778}     &0.746     & \multicolumn{1}{l|}{0.862}    & \multicolumn{1}{l|}{0.881}     &0.863      \\ \hline
			\emph{TrajRL}                    & \multicolumn{1}{l|}{0.742}    & \multicolumn{1}{l|}{0.781}     &0.775     & \multicolumn{1}{l|}{0.876}    & \multicolumn{1}{l|}{0.899}     &0.881      \\ \hline
			\emph{AutoPRL}                 & \multicolumn{1}{l|}{\textbf{0.757}}    & \multicolumn{1}{l|}{\textbf{0.802}}     &\textbf{0.796}      & \multicolumn{1}{l|}{\textbf{0.883}}    & \multicolumn{1}{l|}{\textbf{0.924}}     & \textbf{0.907}     \\ \toprule[1.5pt]
	\end{tabular}}
	\label{tab:NLP}
	\vspace{-10pt}
\end{table}
\noindent

\noindent
\textbf{Performance at Path Imputation. }
The results for path imputation are reported in Table~\ref{tab:NLP}. \emph{AutoPRL} achieves the best observed performance across Recall@3, Recall@5, and MAP on both the Aalborg and Chengdu datasets.
These results indicate that the learned path representations effectively preserve local structural and contextual information that is useful for recovering missing path segments. In particular, the node- and path-level contrastive objectives enable \emph{AutoPRL} to capture both fine-grained local dependencies and global path semantics, resulting in effective representations for path imputation.

\begin{figure*}[!t]
\centering
\includegraphics[scale=0.325]{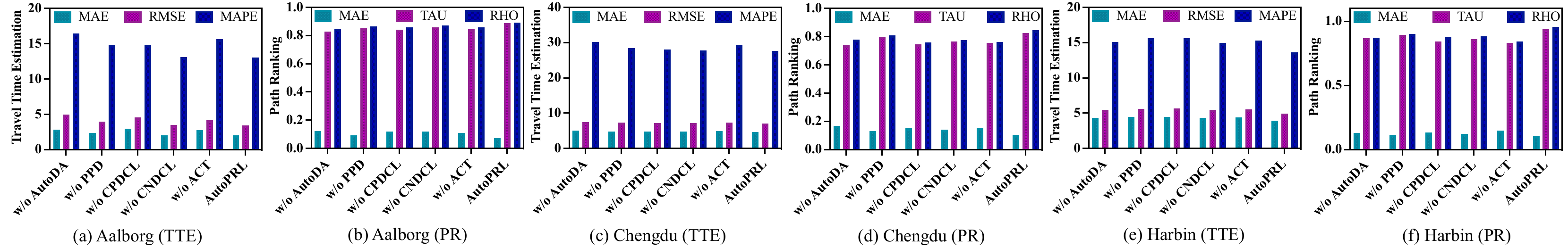}
\caption{Ablation Study.}
\label{fig:AB}
\end{figure*}

\subsection{Effect of Pre-training}
In this experiment, we investigate the effect of pre-training. We employ \emph{AutoPRL} as a pre-training method for the Transformer-based PathRank. We ensure that this \emph{PathRank} framework adopts the same settings as \emph{AutoPRL} for a  fair comparison but the loss terms are changed to mean squared error (MSE) for both tasks. Note that \emph{AutoPRL} is not tied to \emph{PathRank}; the pre-trained path encoder can also initialize other supervised models for task-specific fine-tuning.
In particular, we first train \emph{AutoPRL} in an unsupervised manner, and then use the learned path encoder to initialize the Transformer-based \emph{PathRank}. Specifically, \emph{PathRank} takes path sequences as input and outputs both travel time and the corresponding path ranking score.
Figure \ref{fig:pre-train} reports the performance of \emph{PathRank} with and without pre-training across the two downstream tasks on the three datasets.
When employing both non-pre-trained and pre-trained \emph{PathRank} with labeled data volumes of 2K, 4K, 6K, 8K, 10K, and 12K, 
we observe that (1) with pre-training, \emph{PathRank} consistently outperforms the non-pre-trained version at every labeled data scale. Moreover, \emph{PathRank} with pre-training achieves the same performance as non-pre-trained \emph{PathRank}, but with significantly less labeled data. For example, to match the performance of non-pre-trained \emph{PathRank} trained on 12K labeled samples for path ranking, pre-trained \emph{PathRank} only requires 5K labeled paths for the Aalborg dataset, slightly over 4K for Chengdu, and 8K for Harbin.
(2) \emph{PathRank} with pre-training achieves better performance than the non-pre-trained version, demonstrating its potential as a pre-training method to enhance supervised approaches.

\subsection{Ablation Study }
\label{sec:ab}
\textbf{\emph{AutoPRL} Performance with/without \emph{AutoDA} Module}. 
To evaluate the impact of the automatic view generator, we apply the same augmentation strategies as in \emph{START} to generate path views, resulting in the \emph{w/o AutoDA} model. The results across both tasks and three datasets (Figure \ref{fig:AB}) highlight the effectiveness of the \emph{AutoDA} module. For instance, in Figure \ref{fig:AB}(a), \emph{AutoPRL} improves the MAE from 2.865 to 1.971 for the travel time estimation task on the Aalborg dataset. This substantial reduction in MAE demonstrates the effectiveness of the learned feature-level augmentation compared with predefined augmentation strategies. Furthermore, similar performance gains are consistently observed on the Chengdu and Harbin datasets, reinforcing the general applicability of the \emph{AutoDA} module in diverse real-world scenarios.

\noindent
\textbf{\emph{AutoPRL} Performance with/without Progress-Based Distribution (PPD) for \emph{AutoDA} Module}. 
To assess the impact of the progress-based distribution strategy in training the \emph{AutoDA} module, we use a general fixed prior distribution, resulting in the \emph{w/o PPD} variant. Figure \ref{fig:AB} shows that the progress-based distribution improves the ranking score ($\tau$) from 0.798 to 0.836 on the Chengdu dataset for the path ranking task. This improvement highlights the benefits of adaptively adjusting the distribution of augmented views during training to better match the model’s learning progression. Additionally, similar performance enhancements are observed on the Aalborg and Harbin datasets, further demonstrating that the progress-based strategy consistently improves representation quality and generalization across diverse datasets.

\begin{figure*}[!t]
\centering
\includegraphics[scale=0.33]{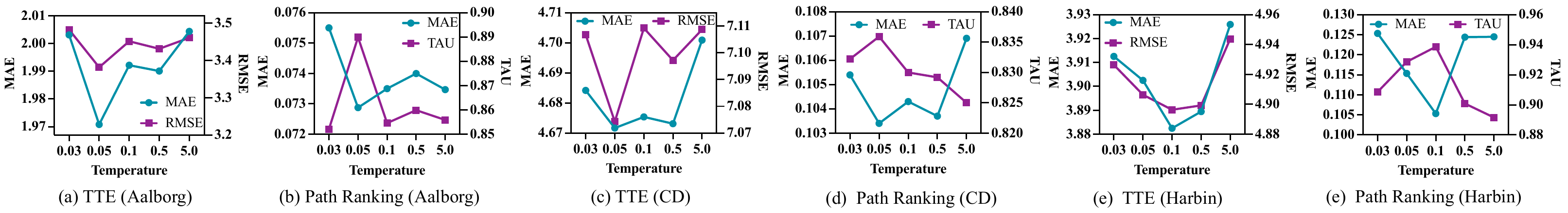}
\caption{Effect of Temperature $t$}
\label{fig:tem}
\end{figure*}

\begin{figure*}[!t]
\centering
\includegraphics[scale=0.3]{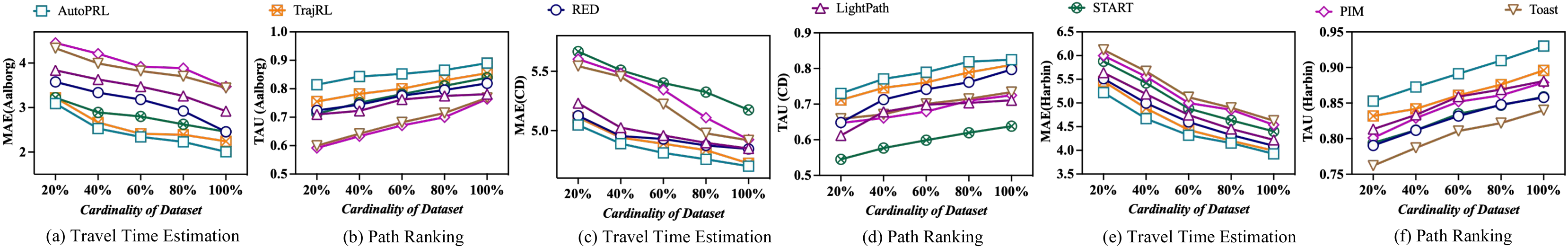}
\caption{Scalability Performance vs. Cardinality}
\label{fig:ScalabilityWC}
\end{figure*}

\begin{figure*}[!t]
\centering
\includegraphics[scale=0.368]{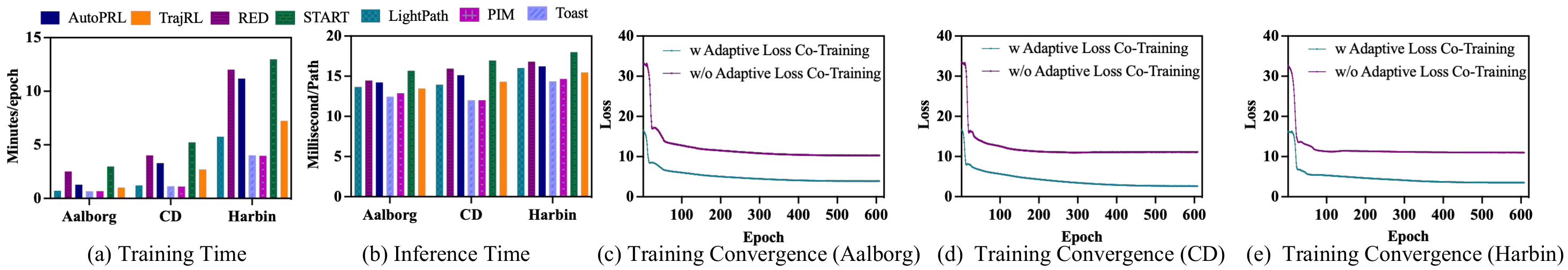}
\caption{(a) Model training time (in minutes) per epoch; (b) model inference time (in ms) per path; and (c) -- (e) training convergence of \emph{AutoPRL} on three datasets.}
\label{fig:TE}
\vspace{-10pt}
\end{figure*}

\noindent
\textbf{\emph{AutoPRL} Performance with/without Cross-View Decoupled Contrastive Loss}. 
To assess the impact of cross-path decoupled contrastive loss (\emph{CPDCL}) and cross-node decoupled contrastive loss (\emph{CNDCL}), we conduct experiments considering each loss independently. (1) With only \emph{CPDCL} (i.e., \emph{w/o CNDCL}), the MAE for \emph{AutoPRL} increases from 3.883 to 4.326 for travel time estimation on the Harbin dataset (Figure \ref{fig:AB}). (2) Conversely, with only \emph{CNDCL} (i.e., \emph{w/o CPDCL}), the MAE increases from 3.883 to 4.460. 
This indicates that \emph{CPDCL} has a larger impact than \emph{CNDCL}, though both contribute to improved \emph{AutoPRL} performance.
These findings suggest that modeling cross-path dependencies captures broader structural relationships within paths, which are essential for accurate travel time estimation. Furthermore, incorporating both \emph{CPDCL} and \emph{CNDCL} ensures a comprehensive understanding of path and node-level information, further boosting the model’s effectiveness.

\noindent
\textbf{\emph{AutoPRL} Performance with/without Multi-loss Optimization}. 
To evaluate the effect of the adaptive loss co-training (ACT) strategy, we train both the view generator and path encoder with fixed loss weights, similar to existing methods that simply sum the losses without adaptive optimization (denoted as \emph{w/o ACT}). Figure \ref{fig:AB} shows the results across both tasks and three datasets, demonstrating that \emph{AutoPRL} with \emph{ACT} achieves superior performance compared to \emph{AutoPRL} without \emph{ACT}, confirming the effectiveness of this strategy. This improvement stems from the ACT module’s ability to dynamically balance the learning of different components, effectively mitigating the dominance of any single loss during training. Additionally, ACT enables better convergence and generalization across datasets with varying characteristics, further enhancing the generality of \emph{AutoPRL}.

\begin{table*}[t]
	\centering
	\caption{Ablation study on travel-time estimation and path ranking under different generative modeling paradigms.}
	{\renewcommand{\arraystretch}{1.3}
		\begin{tabular}{l|llllll|llllll}
			\toprule[1.5pt]
			\multirow{3}{*}{\textbf{Method}} & \multicolumn{6}{l|}{\textbf{Travel Time Estimation}}                                                                                                    & \multicolumn{6}{l}{\textbf{Path Ranking}}                                                                                                          \\ \cline{2-13} 
			& \multicolumn{3}{l|}{\textbf{Aalborg}}                                                     & \multicolumn{3}{l|}{\textbf{Chengdu}}                                & \multicolumn{3}{l|}{\textbf{Aalborg}}                                                   & \multicolumn{3}{l}{\textbf{Chengdu}}                              \\ \cline{2-13} 
			& \multicolumn{1}{l|}{\textbf{MAE}} & \multicolumn{1}{l|}{\textbf{MAPE}} & \multicolumn{1}{l|}{\textbf{RMSE}} & \multicolumn{1}{l|}{\textbf{MAE}} & \multicolumn{1}{l|}{\textbf{MAPE}} & \textbf{RMSE} & \multicolumn{1}{l|}{\textbf{MAE}} & \multicolumn{1}{l|}{\textbf{TAU}($\tau$)} & \multicolumn{1}{l|}{\textbf{RHO}($\rho$)} & \multicolumn{1}{l|}{\textbf{MAE}} & \multicolumn{1}{l|}{\textbf{TAU}($\tau$)} & \textbf{RHO}($\rho$) \\ \toprule[1pt]
			\emph{AutoPRL-DGM}             & \multicolumn{1}{l|}{1.976}    & \multicolumn{1}{l|}{12.901}     & \multicolumn{1}{l|}{3.396}     & \multicolumn{1}{l|}{4.683}    & \multicolumn{1}{l|}{27.447}     &7.091      & \multicolumn{1}{l|}{0.079}    & \multicolumn{1}{l|}{0.870}    & \multicolumn{1}{l|}{0.878}    & \multicolumn{1}{l|}{0.112}    & \multicolumn{1}{l|}{0.821}    &0.830     \\ \hline
			\emph{AutoPRL-GMGP}              & \multicolumn{1}{l|}{\textbf{1.971}}    & \multicolumn{1}{l|}{\textbf{12.867}}     & \multicolumn{1}{l|}{\textbf{3.382}}     & \multicolumn{1}{l|}{\textbf{4.672}}    & \multicolumn{1}{l|}{\textbf{27.318}}     &\textbf{7.074}      & \multicolumn{1}{l|}{\textbf{0.073}}    & \multicolumn{1}{l|}{\textbf{0.890}}    & \multicolumn{1}{l|}{\textbf{0.893}}    & \multicolumn{1}{l|}{\textbf{0.103}}    & \multicolumn{1}{l|}{\textbf{0.836}}    &\textbf{0.845}     \\ \toprule[1.5pt]
	\end{tabular}}
	\label{tab:abnn}
	\vspace{-5pt}
\end{table*}
\noindent
\textbf{\emph{AutoPRL} Performance with Different Generative Modeling Paradigms.}
To verify the effectiveness of the proposed VAE-based view generator, we compare \emph{AutoPRL} with a diffusion-based generative variant, denoted as \emph{AutoPRL-DGM}. As shown in Table~\ref{tab:abnn}, the original VAE-based design achieves
slightly better observed performance than \emph{AutoPRL-DGM} on both
travel-time estimation and path ranking across the Aalborg and Chengdu
datasets. Although the absolute performance differences are relatively
modest, these results suggest that the VNAE-based formulation provides a
stable and effective choice for path view generation in our setting.

\subsection{Parameter Study }
\noindent
\textbf{Effect of Temperature. }Further, we evaluate the sensitivity of \emph{AutoPRL} to assess its performance under varying temperature conditions in the decoupled contrastive loss. Specifically, we conduct a parameter study across three datasets, evaluating model performance for $t \in \{0.03,0.05,0.1,0.5,5\}$. 
The results, presented in Figure \ref{fig:tem}, indicate that the best performance is achieved at $t=0.05$ on the Aalborg and Chengdu datasets, while $t=0.1$ performs best on the Harbin dataset. These findings suggest that optimal temperature values vary across datasets, reflecting their differing complexity levels.
We also observe that when $t=5$, \emph{AutoPRL} exhibits reduced performance, as the contrastive learning component applies less pressure to distinguish between positive and negative samples, leading to collapsed embeddings and diminished discriminative power.
Overall, we conclude that the temperature parameter should be tuned separately for each dataset, as optimal values may vary depending on dataset complexity and characteristics.

\begin{figure}[!t]
	\centering
	\includegraphics[scale=0.3]{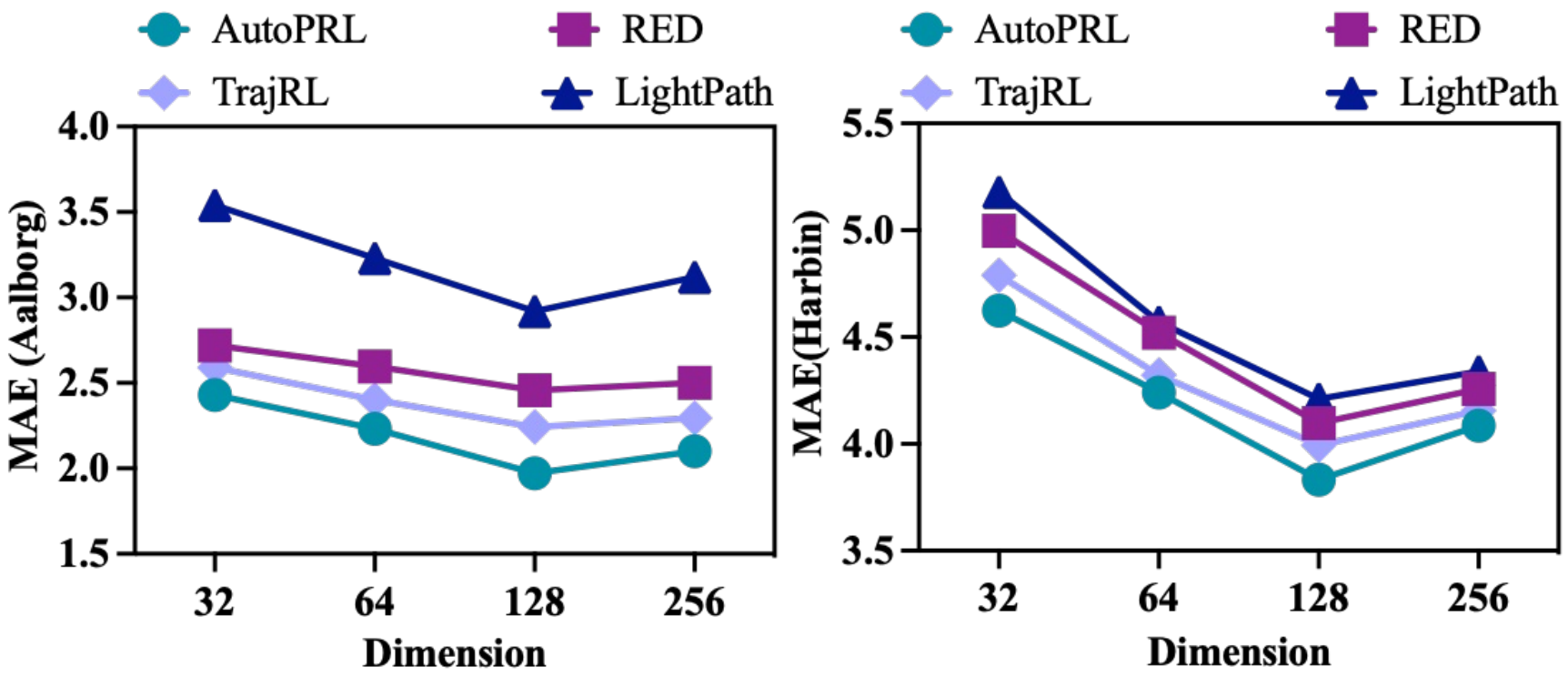}
	\caption{Effect of Embedding Dimension.}
	\label{fig:DIM}
	\vspace{-10pt}
\end{figure}
\noindent
\textbf{Effect of Dimension. }To examine the influence of the path embedding size $d_p$, we conduct a sensitivity analysis on the travel-time estimation task using the Aalborg and Harbin datasets. We compare \emph{AutoPRL} with representative baselines, including \emph{RED}, \emph{TrajRL}, and \emph{LightPath}, under $d_p \in \{32, 64, 128, 256\}$. As shown in Figure~\ref{fig:DIM}, \emph{AutoPRL} consistently achieves the lowest MAE across all embedding sizes on both datasets. In addition, $d_p=128$ generally yields the best or near-best performance, while increasing the dimension to 256 does not further improve accuracy and introduces additional computational cost. Following \emph{RED} and our sensitivity results, we set $d_p=128$ as the default embedding size for all methods to ensure a fair comparison and a good balance between effectiveness and efficiency.

\begin{table*}[t]
	\caption{Batch-size Sensitivity Analysis on Travel-Time Estimation Comparing \emph{AutoPRL-SimCLR} and \emph{AutoPRL-DCLW}.}
	\centering
	{\renewcommand{\arraystretch}{1.3}
		\begin{tabular}{l|llllllll|llllllll}
			\toprule[1.5pt]
			\multirow{3}{*}{\begin{tabular}[c]{@{}l@{}}\textbf{Training}\\ \textbf{Epochs}\end{tabular}} & \multicolumn{8}{l|}{\textbf{Aalborg}}                                                                                                                                                                                                                              & \multicolumn{8}{l}{\textbf{Harbin}}                                                                                                                                                                                                                               \\ \cline{2-17} 
			& \multicolumn{4}{l|}{\textit{\textbf{AutoPRL-SimCLR}}}                                                                                      & \multicolumn{4}{l|}{\textit{\textbf{AutoPRL-DCLW}}}                                                                   & \multicolumn{4}{l|}{\textit{\textbf{AutoPRL-SimCLR}}}                                                                                      & \multicolumn{4}{l}{\textit{\textbf{AutoPRL-DCLW}}}                                                                   \\ \cline{2-17} 
			& \multicolumn{1}{l|}{\textbf{16}} & \multicolumn{1}{l|}{\textbf{32}} & \multicolumn{1}{l|}{\textbf{64}} & \multicolumn{1}{l|}{\textbf{128}} & \multicolumn{1}{l|}{\textbf{16}} & \multicolumn{1}{l|}{\textbf{32}} & \multicolumn{1}{l|}{\textbf{64}} & \textbf{128} & \multicolumn{1}{l|}{\textbf{16}} & \multicolumn{1}{l|}{\textbf{32}} & \multicolumn{1}{l|}{\textbf{64}} & \multicolumn{1}{l|}{\textbf{128}} & \multicolumn{1}{l|}{\textbf{16}} & \multicolumn{1}{l|}{\textbf{32}} & \multicolumn{1}{l|}{\textbf{64}} & \textbf{128} \\ \toprule[1pt]
			\textbf{400}                                                                         & \multicolumn{1}{l|}{2.844}            & \multicolumn{1}{l|}{2.657}            & \multicolumn{1}{l|}{2.539}            & \multicolumn{1}{l|}{2.340}             & \multicolumn{1}{l|}{\textbf{2.112}}            & \multicolumn{1}{l|}{\textbf{2.044}}            & \multicolumn{1}{l|}{\textbf{1.971}}            &\textbf{1.954}              & \multicolumn{1}{l|}{--}            & \multicolumn{1}{l|}{--}            & \multicolumn{1}{l|}{--}            & \multicolumn{1}{l|}{--}             & \multicolumn{1}{l|}{--}            & \multicolumn{1}{l|}{--}            & \multicolumn{1}{l|}{--}            &--              \\ \hline
			\textbf{500}                                                                         & \multicolumn{1}{l|}{\textbf{2.435}}            & \multicolumn{1}{l|}{\textbf{2.216}}            & \multicolumn{1}{l|}{\textbf{2.161}}            & \multicolumn{1}{l|}{\textbf{2.064}}             & \multicolumn{1}{l|}{--}            & \multicolumn{1}{l|}{--}            & \multicolumn{1}{l|}{--}            &--              & \multicolumn{1}{l|}{6.158}            & \multicolumn{1}{l|}{5.725}            & \multicolumn{1}{l|}{5.384}            & \multicolumn{1}{l|}{5.008}             & \multicolumn{1}{l|}{\textbf{4.001}}            & \multicolumn{1}{l|}{\textbf{3.964}}            & \multicolumn{1}{l|}{\textbf{3.883}}            &\textbf{3.775}              \\ \hline
			\textbf{600}                                                                         & \multicolumn{1}{l|}{--}            & \multicolumn{1}{l|}{--}            & \multicolumn{1}{l|}{--}            & \multicolumn{1}{l|}{--}             & \multicolumn{1}{l|}{--}            & \multicolumn{1}{l|}{--}            & \multicolumn{1}{l|}{--}            & --             & \multicolumn{1}{l|}{5.574}            & \multicolumn{1}{l|}{4.977}            & \multicolumn{1}{l|}{4.472}            & \multicolumn{1}{l|}{4.258}             & \multicolumn{1}{l|}{--}            & \multicolumn{1}{l|}{--}            & \multicolumn{1}{l|}{--}            & --             \\ \hline
			\textbf{700}                                                                         & \multicolumn{1}{l|}{--}            & \multicolumn{1}{l|}{--}            & \multicolumn{1}{l|}{--}            & \multicolumn{1}{l|}{--}             & \multicolumn{1}{l|}{--}            & \multicolumn{1}{l|}{--}            & \multicolumn{1}{l|}{--}            &--              & \multicolumn{1}{l|}{\textbf{4.476}}            & \multicolumn{1}{l|}{\textbf{4.433}}            & \multicolumn{1}{l|}{\textbf{4.212}}            & \multicolumn{1}{l|}{\textbf{3.979}}             & \multicolumn{1}{l|}{--}            & \multicolumn{1}{l|}{--}            & \multicolumn{1}{l|}{--}            &--             \\ \toprule[1.5pt]
	\end{tabular}}
	\label{tab:simvsdcl1}
	\vspace{-10pt}
\end{table*}

\begin{table*}[t]
	\centering
	\caption{Performance on Travel-Time Estimation and Path Ranking under Noisy Representations.}
	{\renewcommand{\arraystretch}{1.3}
		\begin{tabular}{l|llllll|llllll}
			\toprule[1.5pt]
			\multirow{3}{*}{\textbf{Method}} & \multicolumn{6}{l|}{\textbf{Travel Time Estimation}}                                                                                                    & \multicolumn{6}{l}{\textbf{Path Ranking}}                                                                                                          \\ \cline{2-13} 
			& \multicolumn{3}{l|}{\textbf{Aalborg}}                                                     & \multicolumn{3}{l|}{\textbf{Chengdu}}                                & \multicolumn{3}{l|}{\textbf{Aalborg}}                                                   & \multicolumn{3}{l}{\textbf{Chengdu}}                              \\ \cline{2-13} 
			& \multicolumn{1}{l|}{\textbf{MAE}} & \multicolumn{1}{l|}{\textbf{MAPE}} & \multicolumn{1}{l|}{\textbf{RMSE}} & \multicolumn{1}{l|}{\textbf{MAE}} & \multicolumn{1}{l|}{\textbf{MAPE}} & \textbf{RMSE} & \multicolumn{1}{l|}{\textbf{MAE}} & \multicolumn{1}{l|}{\textbf{TAU}($\tau$)} & \multicolumn{1}{l|}{\textbf{RHO}($\rho$)} & \multicolumn{1}{l|}{\textbf{MAE}} & \multicolumn{1}{l|}{\textbf{TAU}($\tau$)} & \textbf{RHO}($\rho$) \\ \toprule[1pt]

			\emph{Toast}             & \multicolumn{1}{l|}{3.439}    & \multicolumn{1}{l|}{21.311}     & \multicolumn{1}{l|}{5.555}     & \multicolumn{1}{l|}{4.918}    & \multicolumn{1}{l|}{28.567}     &7.513      & \multicolumn{1}{l|}{0.138}    & \multicolumn{1}{l|}{0.768}    & \multicolumn{1}{l|}{0.779}    & \multicolumn{1}{l|}{0.131}    & \multicolumn{1}{l|}{0.734}    &0.752     \\ \hline

			\emph{Toast w/ noise}             & \multicolumn{1}{l|}{4.426}    & \multicolumn{1}{l|}{21.906}     & \multicolumn{1}{l|}{6.066}     & \multicolumn{1}{l|}{5.622}    & \multicolumn{1}{l|}{29.134}     &8.233      & \multicolumn{1}{l|}{0.207}    & \multicolumn{1}{l|}{0.672}    & \multicolumn{1}{l|}{0.683}    & \multicolumn{1}{l|}{0.206}    & \multicolumn{1}{l|}{0.692}    &0.703     \\ \hline
			
			\emph{PIM}             & \multicolumn{1}{l|}{3.471}    & \multicolumn{1}{l|}{21.314}     & \multicolumn{1}{l|}{5.640}     & \multicolumn{1}{l|}{4.924}    & \multicolumn{1}{l|}{28.553}     &7.493      & \multicolumn{1}{l|}{0.141}    & \multicolumn{1}{l|}{0.764}    & \multicolumn{1}{l|}{0.777}    & \multicolumn{1}{l|}{0.125}    & \multicolumn{1}{l|}{0.725}    &0.745     \\ \hline

			\emph{PIM w/ noise}             & \multicolumn{1}{l|}{4.003}    & \multicolumn{1}{l|}{21.904}     & \multicolumn{1}{l|}{5.982}     & \multicolumn{1}{l|}{5.427}    & \multicolumn{1}{l|}{28.994}     &7.89      & \multicolumn{1}{l|}{0.192}    & \multicolumn{1}{l|}{0.727}    & \multicolumn{1}{l|}{0.738}    & \multicolumn{1}{l|}{0.169}    & \multicolumn{1}{l|}{0.674}    &0.685     \\ \hline
			
			\emph{START}             & \multicolumn{1}{l|}{2.965}    & \multicolumn{1}{l|}{18.432}     & \multicolumn{1}{l|}{4.979}     & \multicolumn{1}{l|}{5.174}    & \multicolumn{1}{l|}{30.769}     &7.726      & \multicolumn{1}{l|}{0.178}    & \multicolumn{1}{l|}{0.839}    & \multicolumn{1}{l|}{0.859}    & \multicolumn{1}{l|}{0.228}    & \multicolumn{1}{l|}{0.638}    &0.679     \\ \hline

			\emph{START w/ noise}             & \multicolumn{1}{l|}{3.441}    & \multicolumn{1}{l|}{18.901}     & \multicolumn{1}{l|}{5.551}     & \multicolumn{1}{l|}{5.607}    & \multicolumn{1}{l|}{31.114}     &8.011      & \multicolumn{1}{l|}{0.225}    & \multicolumn{1}{l|}{0.795}    & \multicolumn{1}{l|}{0.803}    & \multicolumn{1}{l|}{0.265}    & \multicolumn{1}{l|}{0.603}    &0.617     \\ \hline
			
			\emph{LightPath}             & \multicolumn{1}{l|}{2.920}    & \multicolumn{1}{l|}{18.254}     & \multicolumn{1}{l|}{4.932}     & \multicolumn{1}{l|}{4.850}    & \multicolumn{1}{l|}{28.387}     &7.406      & \multicolumn{1}{l|}{0.118}    & \multicolumn{1}{l|}{0.781}    & \multicolumn{1}{l|}{0.791}    & \multicolumn{1}{l|}{0.195}    & \multicolumn{1}{l|}{0.711}    &0.748     \\ \hline
			
			\emph{LightPath w/ noise}             & \multicolumn{1}{l|}{3.232}    & \multicolumn{1}{l|}{18.673}     & \multicolumn{1}{l|}{5.412}     & \multicolumn{1}{l|}{5.365}    & \multicolumn{1}{l|}{28.387}     &7.798      & \multicolumn{1}{l|}{0.156}    & \multicolumn{1}{l|}{0.736}    & \multicolumn{1}{l|}{0.744}    & \multicolumn{1}{l|}{0.248}    & \multicolumn{1}{l|}{0.672}    &0.681     \\ \hline
			
			\emph{RED}             & \multicolumn{1}{l|}{2.457}    & \multicolumn{1}{l|}{14.907}     & \multicolumn{1}{l|}{3.809}     & \multicolumn{1}{l|}{4.845}    & \multicolumn{1}{l|}{28.380}     &7.296      & \multicolumn{1}{l|}{0.109}    & \multicolumn{1}{l|}{0.819}    & \multicolumn{1}{l|}{0.823}    & \multicolumn{1}{l|}{0.129}    & \multicolumn{1}{l|}{0.797}    &0.833     \\ \hline
			
			\emph{RED w/ noise}             & \multicolumn{1}{l|}{2.527}    & \multicolumn{1}{l|}{14.999}     & \multicolumn{1}{l|}{3.903}     & \multicolumn{1}{l|}{4.987}    & \multicolumn{1}{l|}{28.510}     &7.374      & \multicolumn{1}{l|}{0.127}    & \multicolumn{1}{l|}{0.790}    & \multicolumn{1}{l|}{0.805}    & \multicolumn{1}{l|}{0.144}    & \multicolumn{1}{l|}{0.762}    &0.779     \\ \hline

			\emph{TrajRL}              & \multicolumn{1}{l|}{2.243}             & \multicolumn{1}{l|}{13.125}              &\multicolumn{1}{l|}{3.632}               & \multicolumn{1}{l|}{4.724}             & \multicolumn{1}{l|}{27.960}              &\multicolumn{1}{l|}{7.108}       & \multicolumn{1}{l|}{0.089}    & \multicolumn{1}{l|}{0.855}    & \multicolumn{1}{l|}{0.871}    & \multicolumn{1}{l|}{0.114}    & \multicolumn{1}{l|}{0.810}    &0.839    \\ \hline

			\emph{TrajRL w/ noise}              & \multicolumn{1}{l|}{2.357}    & \multicolumn{1}{l|}{13.396}     & \multicolumn{1}{l|}{3.829}     & \multicolumn{1}{l|}{4.810}    & \multicolumn{1}{l|}{28.128}     &7.278      & \multicolumn{1}{l|}{0.097}    & \multicolumn{1}{l|}{0.823}    & \multicolumn{1}{l|}{0.841}    & \multicolumn{1}{l|}{0.128}    & \multicolumn{1}{l|}{0.782}    &0.820     \\ \hline
			
			\emph{AutoPRL}              & \multicolumn{1}{l|}{1.971}    & \multicolumn{1}{l|}{12.867}     & \multicolumn{1}{l|}{3.382}     & \multicolumn{1}{l|}{4.672}    & \multicolumn{1}{l|}{27.318}     &7.074      & \multicolumn{1}{l|}{0.073}    & \multicolumn{1}{l|}{0.890}    & \multicolumn{1}{l|}{0.893}    & \multicolumn{1}{l|}{0.103}    & \multicolumn{1}{l|}{0.836}    &0.845     \\ \hline
			
			\emph{AutoPRL w/ noise}              & \multicolumn{1}{l|}{1.985}    & \multicolumn{1}{l|}{12.875}     & \multicolumn{1}{l|}{3.392}     & \multicolumn{1}{l|}{4.690}    & \multicolumn{1}{l|}{27.338}     &7.094      & \multicolumn{1}{l|}{0.078}    & \multicolumn{1}{l|}{0.881}    & \multicolumn{1}{l|}{0.890}    & \multicolumn{1}{l|}{0.106}    & \multicolumn{1}{l|}{0.833}    &0.840     \\ \toprule[1.5pt]
	\end{tabular}}
	\label{tab:noisy}
	\vspace{-10pt}
\end{table*}
\subsection{Efficiency Study}
We then report the model evaluations in terms of scalability and efficiency.

\noindent
\textbf{Model Scalability. }In this section, we explore the scalability (i.e., training data size) performance of \emph{AutoPRL} in comparison with \emph{LightPath} and \emph{RED}, using the Aalborg, Chengdu (CD), and Harbin datasets. Figure \ref{fig:ScalabilityWC} illustrates the experimental results, where 20\%, 40\%, 60\%, 80\%, and 100\% represent the proportion of the training dataset used. 
The first observation is that performance improves with increasing data cardinality, as larger training datasets enhance model capabilities.
Second, \emph{AutoPRL} demonstrates consistent performance improvements over \emph{Toast}, \emph{PIM}, \emph{LightPath}, \emph{START}, \emph{RED} and \emph{TrajRL}. Moreover, the performance of \emph{AutoPRL} in both travel time estimation and path ranking tasks across the Aalborg, Chengdu, and Harbin datasets shows a consistent improvement as cardinality increases. %
Overall, our \emph{AutoPRL} model demonstrates strong potential for scalability, supporting large-scale applications across diverse datasets and tasks.

\noindent
\textbf{Runtime Efficiency. }In this section, we investigate the runtime efficiency during both the training and inference phases. Figures \ref{fig:TE}(a) and \ref{fig:TE}(b) illustrate the corresponding results. Overall, \emph{AutoPRL} achieves competitive runtime efficiency compared with the baselines. Specifically, \emph{AutoPRL} is faster than \emph{RED} in both training and inference phases across all datasets, with more evident improvements on Aalborg and Chengdu during training. Although lightweight methods such as \emph{LightPath} can achieve lower runtime costs due to their simpler input processing strategies, \emph{AutoPRL} maintains comparable efficiency while delivering strong overall performance.
During inference, we measure the inference time for each path embedding. As observed, \emph{AutoPRL} and \emph{LightPath} require approximately the same amount of time per path embedding, but are slightly faster than \emph{RED}.

\noindent
\textbf{Training Efficiency. }To further evaluate the efficiency of \emph{AutoPRL}, we examine its convergence behavior with and without the \emph{ACT} strategy (refer to Section \ref{sec:ab}) on the three datasets. Figures \ref{fig:TE}(c), \ref{fig:TE}(d), and \ref{fig:TE}(e) present the corresponding results. The number of epochs for both \emph{w/ ACT} and \emph{w/o ACT} is set to 600. We observe the following: (1) both models quickly reduce their loss during the initial epochs; and (2) \emph{AutoPRL} \emph{w/ ACT} consistently maintains a lower loss than \emph{AutoPRL} \emph{w/o ACT}, indicating that \emph{AutoPRL} achieves superior training efficiency, converging faster and more stably to better performance with reduced computational cost. This improvement can be attributed to the adaptive optimization of multiple losses under the \emph{ACT} strategy, which facilitates effective coordination between the view generator and path encoder. 

\begin{figure*}[!t]
	\centering
	\includegraphics[scale=0.25]{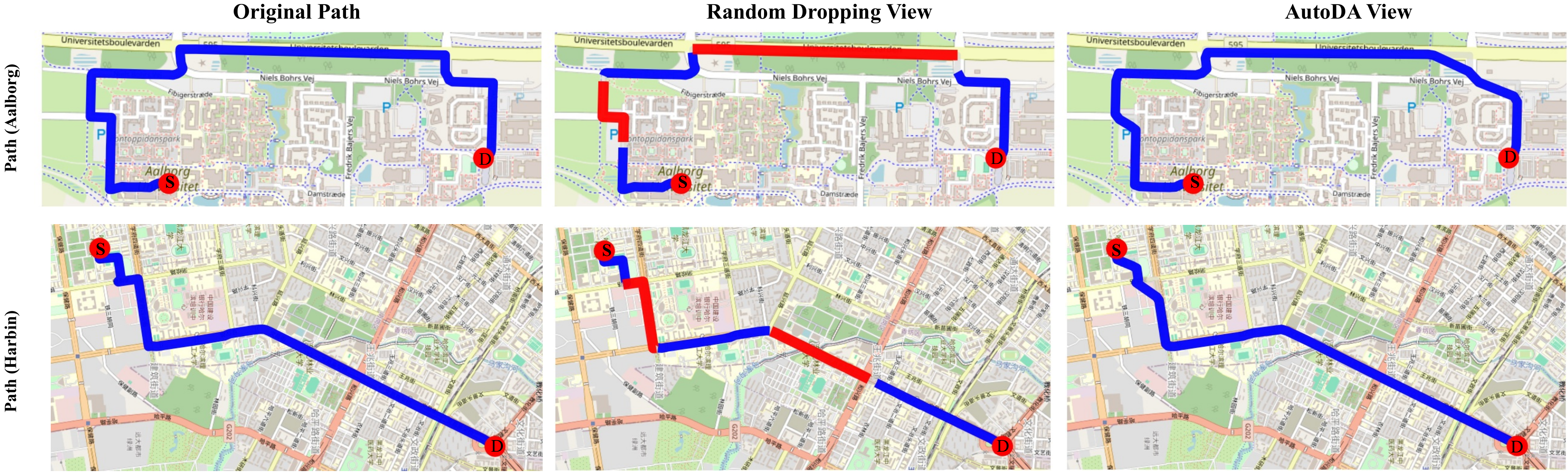}
	\caption{Case study of path view generation. We compare the original path, a random dropping view, and an \emph{AutoDA}-generated view that is projected onto the road network only for visualization. Random dropping may remove important intermediate nodes and break local path continuity, while \emph{AutoDA} better preserves the overall route semantics.}
	\label{fig:VIS}
	\vspace{-10pt}
\end{figure*}

\noindent
\textbf{Batch-Size Sensitivity Analysis.}
To evaluate the sensitivity of \emph{CVDC} to batch size, we compare \emph{AutoPRL-DCLW} with a standard SimCLR-based variant, \emph{AutoPRL-SimCLR}, under batch sizes of 16, 32, 64, and 128 on the travel-time estimation task using the Aalborg and Harbin datasets. As shown in Table~\ref{tab:simvsdcl1}, \emph{AutoPRL-DCLW} consistently achieves lower MAE than \emph{AutoPRL-SimCLR} across different batch sizes.
Moreover, \emph{AutoPRL-DCLW} converges within 400 epochs on Aalborg and 500 epochs on Harbin, whereas \emph{AutoPRL-SimCLR} requires 500 and 700 epochs, respectively. These results demonstrate that \emph{CVDC} is less sensitive to batch size and enables more stable and efficient training than the standard SimCLR objective.
\subsection{Robustness to Noisy Representations}

To evaluate the robustness of the learned path representations, we perturb the
learned path embedding $\mathbf{h}_p$ by adding Gaussian noise:
\begin{equation}
	\tilde{\mathbf{h}}_p
	=
	\mathbf{h}_p+\boldsymbol{\epsilon},
	\qquad
	\boldsymbol{\epsilon}
	\sim
	\mathcal{N}
	\left(
	\mathbf{0},
	\sigma_{\mathrm{noise}}^{2}I
	\right),
\end{equation}
where $\sigma_{\mathrm{noise}}=0.03$ controls the perturbation magnitude.
The noise is independently sampled for each embedding dimension and is applied
to the learned path representations during downstream evaluation, while the
path encoder and downstream prediction settings remain unchanged. We use the
same noise magnitude and evaluation protocol for all compared methods to ensure
a controlled comparison.

As reported in Table~\ref{tab:noisy}, \emph{AutoPRL} exhibits only marginal
performance degradation under noisy representations. For example, the MAE on
Aalborg travel-time estimation increases slightly from 1.971 to 1.985, whereas
most baseline methods experience more noticeable performance drops.

The results indicate that the path representations learned by \emph{AutoPRL}
are relatively robust to representation-level perturbations. Together with the
results from the preceding experiments, this provides additional empirical
evidence that \emph{AutoPRL} can learn stable and informative path
representations under moderate perturbations.

\subsection{Visualization of Generated Path Views}

To provide more direct evidence for the semantic consistency of the generated views, we conduct a qualitative case study of path view generation. Figure~\ref{fig:VIS} compares the original path, a random dropping view, and an \emph{AutoDA}-generated view projected onto the road network only for visualization. Random dropping may remove important intermediate nodes and break local path continuity, leading to a view that deviates from the original route semantics. In contrast, the \emph{AutoDA}-generated view better preserves the overall route semantics while still introducing moderate variation.

It is worth noting that the projection in Figure~\ref{fig:VIS} is used only for visualization. During training, \emph{AutoDA} operates on continuous topology-aware node-embedding sequences in the latent space and does not require mapping the generated views back to discrete road-network nodes.

\section{Conclusion}
We propose \emph{AutoPRL}, an effective and efficient framework for path representation learning that leverages decoupled contrastive learning. Unlike existing contrastive learning methods that rely on pre-defined augmentation strategies and often suffer from information loss, \emph{AutoPRL} first introduces a learnable path view generator parameterized by Gaussian distributions to automatically create diverse, semantically meaningful views. To improve efficiency, we design cross-view decoupled contrastive strategies that disentangle positive and negative pair optimization at both node and path levels, and integrate an Adaptive Co-Training (ACT) module that dynamically assigns uncertainty-based weights to each objective. Extensive experiments across multiple downstream tasks on three real-world datasets demonstrate that \emph{AutoPRL} achieves strong performance and consistent empirical improvements over the considered baselines.

\section*{Acknowledgements}
This work was partially supported by the National Natural Science Foundation of China (Nos. 62472174 and 62402082), the ECNU Multifunctional Platform for Innovation (No. 001), and the Chongqing Municipal Education Commission (No. KJQN202400637). 

\balance
\bibliographystyle{IEEEtran}
\bibliography{IEEEexample}
\end{document}